\documentclass[article,onefignum,onetabnum]{siamonline220329}

\usepackage{lipsum}
\usepackage{amsfonts}
\usepackage{graphicx}
\usepackage{epstopdf}
\usepackage{caption}
\usepackage{subcaption}
\usepackage{placeins}
\usepackage{multicol}
\usepackage{tabularx}
\usepackage{multirow}
\usepackage{soul}
\usepackage{amsmath}
\usepackage{wrapfig}
\usepackage{bm}
\usepackage{algorithm, algorithmicx, algpseudocode}
\ifpdf
  \DeclareGraphicsExtensions{.eps,.pdf,.png,.jpg}
\else
  \DeclareGraphicsExtensions{.eps}
\fi

\usepackage{enumitem}
\setlist[enumerate]{leftmargin=.5in}
\setlist[itemize]{leftmargin=.5in}

\newsiamremark{remark}{Remark}
\newsiamremark{hypothesis}{Hypothesis}
\crefname{hypothesis}{Hypothesis}{Hypotheses}
\newsiamthm{claim}{Claim}

\headers{Fast \& Fair}{A. J. Minch, H. A. Vu, and A. Warren}

\title{Fast \& Fair: Efficient Second-Order Robust  Optimization for Fairness in Machine Learning\thanks{Submitted to the editors DATE.
\funding{This work is supported in part by the US NSF award DMS-2051019.}}}

\author{
Allen Joseph Minch\thanks{Department of Mathematics, Brandeis University, Waltham, MA (\email{allenminch@brandeis.edu})}
\and Hung Anh Dinh Vu\thanks{Department of Mathematics, University of Maryland, College Park, MD 
  (\email{hvu1@terpmail.umd.edu})}
\and Anne Marie Warren\thanks{University of Minnesota, Minneapolis, MN (\email{warre659@umn.edu}, \url{https://anniewarren.github.io})}
}

\dedication{\small\textit{Project Advisor}: Dr.~Elizabeth Newman\thanks{Department of Mathematics, Emory University, Atlanta, GA (\email{elizabeth.newman@emory.edu}).}
}

\usepackage{amsopn}

\usepackage{amsmath, amsfonts} 
\usepackage{tikz} 
\usetikzlibrary{shapes.arrows, arrows, decorations.markings}
\usetikzlibrary{calc}
\usepackage{graphicx} 
\usepackage{xcolor} 
\usepackage{multicol} 
\usepackage{hyperref} 
\usepackage{framed}

\DeclareMathOperator*{\argmin}{argmin}
\DeclareMathOperator*{\subjectto}{s.t.}
\DeclareMathOperator*{\argmax}{argmax}

\def\mydefb#1{\expandafter\def\csname bf#1\endcsname{\mathbf{#1}}}
\def\mydefallb#1{\ifx#1\mydefallb\else\mydefb#1\expandafter\mydefallb\fi}
\mydefallb aAbBcCdDeEfFgGhHiIjJkKlLmMnNoOpPqQrRsStTuUvVwWxXyYzZ\mydefallb

\def\mydefb#1{\expandafter\def\csname hat#1\endcsname{\hat{#1}}}
\def\mydefallb#1{\ifx#1\mydefallb\else\mydefb#1\expandafter\mydefallb\fi}
\mydefallb aAbBcCdDeEfFgGhHiIjJkKlLmMnNoOpPqQrRsStTuUvVwWxXyYzZ\mydefallb

\def\mydefb#1{\expandafter\def\csname bfhat#1\endcsname{\hat{\mathbf{#1}}}}
\def\mydefallb#1{\ifx#1\mydefallb\else\mydefb#1\expandafter\mydefallb\fi}
\mydefallb aAbBcCdDeEfFgGhHiIjJkKlLmMnNoOpPqQrRsStTuUvVwWxXyYzZ\mydefallb

\def\mydefb#1{\expandafter\def\csname til#1\endcsname{\tilde{#1}}}
\def\mydefallb#1{\ifx#1\mydefallb\else\mydefb#1\expandafter\mydefallb\fi}
\mydefallb aAbBcCdDeEfFgGhHiIjJkKlLmMnNoOpPqQrRsStTuUvVwWxXyYzZ\mydefallb

\def\mydefb#1{\expandafter\def\csname tbf#1\endcsname{\tilde{\mathbf{#1}}}}
\def\mydefallb#1{\ifx#1\mydefallb\else\mydefb#1\expandafter\mydefallb\fi}
\mydefallb aAbBcCdDeEfFgGhHiIjJkKlLmMnNoOpPqQrRsStTuUvVwWxXyYzZ\mydefallb

\def\mydefb#1{\expandafter\def\csname #1bb\endcsname{\mathbb{#1}}}
\def\mydefallb#1{\ifx#1\mydefallb\else\mydefb#1\expandafter\mydefallb\fi}
\mydefallb aAbBcCdDeEfFgGhHiIjJkKlLmMnNoOpPqQrRsStTuUvVwWxXyYzZ\mydefallb

\def\mydefb#1{\expandafter\def\csname #1cal\endcsname{\mathcal{#1}}}
\def\mydefallb#1{\ifx#1\mydefallb\else\mydefb#1\expandafter\mydefallb\fi}
\mydefallb aAbBcCdDeEfFgGhHiIjJkKlLmMnNoOpPqQrRsStTuUvVwWxXyYzZ\mydefallb

\def\mydefb#1{\expandafter\def\csname T#1\endcsname{\boldsymbol{\mathcal{#1}}}}
\def\mydefallb#1{\ifx#1\mydefallb\else\mydefb#1\expandafter\mydefallb\fi}
\mydefallb aAbBcCdDeEfFgGhHiIjJkKlLmMnNoOpPqQrRsStTuUvVwWxXyYzZ\mydefallb

\def\mydefgreek#1{\expandafter\def\csname bf#1\endcsname{\text{\boldmath$\mathbf{\csname #1\endcsname}$}}}
\def\mydefallgreek#1{\ifx\mydefallgreek#1\else\mydefgreek{#1}%
   \lowercase{\mydefgreek{#1}}\expandafter\mydefallgreek\fi}
\mydefallgreek {alpha}{Alpha}{beta}{Beta}{gamma}{Gamma}{delta}{Delta}{epsilon}{Epsilon}{zeta}{Zeta}{eta}{Eta}{theta}{Theta}{iota}{Iota}{kappa}{Kappa}{lambda}{Lambda}{mu}{Mu}{nu}{Nu}{omicron}{Omicron}{pi}{Pi}{rho}{Rho}{sigma}{Sigma}{tau}{Tau}{upsilon}{Upsilon}{phi}{Phi}{xi}{Xi}{chi}{Chi}{psi}{Psi}{omega}{Omega}\mydefallgreek

\ifpdf
\hypersetup{
  pdftitle={Fast \& Fair},
  pdfauthor={A. Minch, H. A. Vu, and A. M. Warren}
}
\fi

\graphicspath{{./}{Figures/}}

\begin{document}

\maketitle

\begin{abstract}
This project explores adversarial training techniques to develop fairer Deep Neural Networks (DNNs) to mitigate the inherent bias they are known to exhibit. DNNs are susceptible to inheriting bias with respect to sensitive attributes such as race and gender, which can lead to life-altering outcomes (e.g., demographic bias in facial recognition software used to arrest a suspect). We propose a robust optimization problem, which we demonstrate can improve fairness in several datasets, both synthetic and real-world, using an affine linear model. Leveraging second order information, we are able to find a solution to our optimization problem more efficiently than a purely first order method.
\end{abstract}



\section{Introduction}
\label{sec:introduction}
Machine learning has become an integral part of data analysis with its powerful ability to reveal underlying patterns and structures in data. Deep Neural Networks (DNNs) in particular are the gold standard classifying complex data; however, there is a tendency for DNNs to inherit bias from the datasets on which they train. Bias in this sense is not statistical bias, but the ways in which individual advantages or disadvantages manifest in data. This can be especially problematic in areas where machine learning is used to make life-altering decisions such as criminal justice \cite{FURL2002797} and corporate hiring \cite{DBLP:journals/corr/abs-1908-09635}.

The ever-expanding use of machine learning poses a significant ethical question when models are known to perpetuate societal biases \cite{DBLP:journals/corr/abs-1908-09635}. While an in-depth discussion of these ethical concerns is beyond the scope of this work, they motivate our efforts to improve fairness within the models themselves. The unfortunate truth is that the harmful biases we see in our models and in our data come from deep-rooted societal structures that are at present beyond the abilities of machine learning to correct. However, we feel that in the face of these larger issues it is our duty within our means to work towards fairer outcomes. 

One way to potentially achieve fairer outcomes is to use adversarial training to introduce robustness to a model. Robust optimization aims to make the model less susceptible to small variations in data, known as adversarial attacks, but in doing so decreases model accuracy. Recently, the Fair-Robust-Learning framework was proposed to reduce this unfairness problem in adversarial training \cite{xu2021robust}. The authors demonstrated that a combination of fairness and adversarial regularization yielded fairer models on benchmark image classification datasets.

Our research shares the goal of addressing fairness issues in DNNs through the use of adversarial training techniques, but focuses on an additive bias rather than out-of-distribution bias or other forms. We define fairness on different metrics (independence, separation, and sufficiency vs. average and worst-class boundary, robust, and standard errors) to measure additive bias with respect to sensitive 'hidden' attributes. Without aiming to cater to specific types of data, we explore the effects of adversarial training on this definition of fairness. A simultaneous focus is to improve the efficiency of solving robust optimization problems. To this end, we use second-order information to accelerate training, a concept that was not addressed in previous work \cite{xu2021robust}. 

We implement second-order method, termed the ``trust region subproblem'' (TRS), designed explicitly to address inner optimization challenges encountered when introducing robus training. Our experiments, spanning both synthetic and real-world datasets, demonstrate the capabilities of robust optimization in enhancing fairness. We employ three distinct optimizers for these tests, allowing us to compare their performance. Notably, the integration of \texttt{hessQuik} \cite{Newman2022}, has proven instrumental in efficiently deriving exact Hessians. This approach surpasses the projected gradient descent (PGD) method in terms of time efficiency while producing the same solution. For transparency and further community engagement, we've made our Python implementation, including all our experiments, available on our GitHub repository at Fast-N-Fair (\url{https://github.com/elizabethnewman/fast-n-fair}).

The paper is organized as follows: \Cref{sec:background} introduces DNNs and the necessary notation, robust optimization, and our choice of fairness metrics. \Cref{sec:our_appraoch} describes our proposed second order method, our implementation, and is followed by an analysis of the error produced by approximations used in our methods (\Cref{sec:analysis}). In \Cref{sec:Other Methods}, we introduce alternate methods of solving the robust optimization problem that are implemented as a comparison to our proposed approach. \Cref{sec:numerical} first describes the setup of a synthetic dataset along with the preliminary fairness results, and then extends the discussion to several real world datasets. We also examine the relative computational efficiency of our different methods of performing robust optimization. Lastly, \Cref{sec:conclusion} concludes the paper and discusses potential future work. 

\section{Background} First we must discuss DNNs and some necessary notation, robust optimization, and our choice of fairness metrics.
\label{sec:background}

    \subsection{Notation}
    \label{sec:notation}
    Deep neural networks (DNNs) can be represented by a parameterized mapping $f_{\bftheta}: \Xcal \times \Theta \to \Ycal$ from input-target pairs $(\bfx, \bfy) \in \Dcal$, where $\Dcal \subseteq \Xcal \times \Ycal$ is the data space, $\Xcal \subseteq \Rbb^{n_\text{in}}$ is the input space, and $\Ycal \subseteq \Rbb^{n_\text{out}}$ is the target space, and $\Theta \subset \Rbb^{n_{\bftheta}}$ is the parameter space. 
Our goal is to learn the weights $\bftheta \in \Theta$ such that $f_{\bftheta}(\bfx) \approx \bfy$ for all input-target pairs.  
Typically, learning the weights is posed as the optimization problem 
    \begin{equation}
        \min_{\bftheta} \frac{1}{|\Tcal|}\sum_{(\bfx, \bfy) \in \Tcal} L(f_{\bftheta}(\bfx), \bfy) + R(\bftheta)
        \label{eq:bg-nonrobustOptimization}
    \end{equation}
where $\Tcal \subset \Dcal$ is the training set and $R: \Theta \to \Rbb$ is a regularization term to enforce desirable properties on the weights.\par
For many problems with a well-chosen optimizer, we can solve \eqref{eq:bg-nonrobustOptimization} well. 
However, this can lead to problems such as overfitting, where the model fits the training data well but does not generalize to unseen data, or a lack of robustness, where small changes to the data result in significantly different results (e.g., incorrect classifications). 
    
    \subsection{Robust Optimization}
    \label{sec:robust_optimization}
    \begin{figure}[t]
\centering
\includegraphics[width=0.5\linewidth]{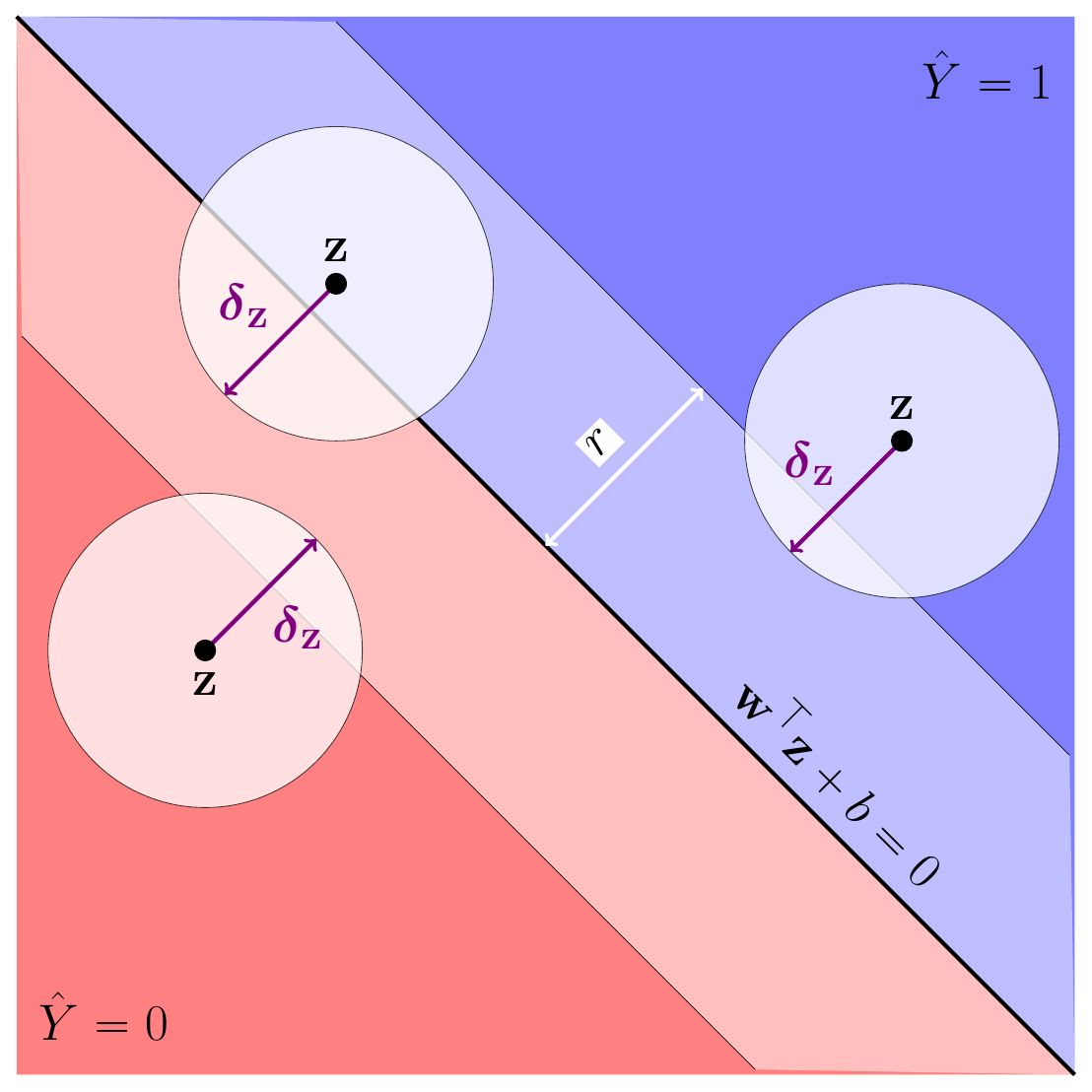} 
            \caption{Robust optimization, visualized in the case of a linear classifier (black line) in two dimensions $\bfw^\top \bfz + b = 0$. 
            The black data points $\bfz \equiv f_{\bftheta}(\bfx)$ are the network outputs for various data inputs. 
            The white circles indicate output features within a radius of $r$ of the network outputs. 
            The direction of perturbation $\bfdelta_{\bfz}$ that maximizes the inner optimization problem is normal to the linear classifier defined by $\bfw$. 
            Any network outputs in the white channel, $r$ away from the linear classifier, change the predicted class. 
            Robust optimization encourages network outputs to live outside of the white channel to avoid ambiguous class predictions.}
            \label{fig:robust}
\end{figure} 
Adversarial training promotes robustness in DNNs by introducing a perturbation $\bfdelta_{\bfx}$ for each input $\bfx$ and solving the minimax problem
    \begin{subequations}\label{eq:bg-robustOptimization}
    \begin{align}
        &\min_{\bftheta}\quad  \frac{1}{|\Tcal|}\sum_{(\bfx, \bfy) \in \Tcal} L(f_{\bftheta}(\bfx + \delta_{\bfx}(\bftheta)),\bfy) + R(\bftheta) \label{eq:bg-outerOptimization}\\
        \subjectto &\quad \bfdelta_{\bfx}(\bftheta) \in \argmax_{\|\bfdelta_{\bfx}\|_2 \le r} L(f_{\bftheta}(\bfx + \bfdelta_{\bfx}), \bfy) \qquad \text{for each $(\bfx,\bfy)\in \Tcal$} \label{eq:bg-innerOptimization}
    \end{align} 
    \end{subequations}
We perturb the inputs $\bfx$ by $\bfdelta_\bfx$ and maximize the Euclidean norm of the perturbation 
$\|\bfdelta_\bfx\|_2$  (inner optimization problem~\eqref{eq:bg-innerOptimization}) while optimally fitting the data (outer minimization problem~\eqref{eq:bg-outerOptimization}). 
We build neighborhoods of radius $r$ around our training points where we can rely on our model classifying anything within the neighborhood similarly. See \cref{fig:robust} for a visualization. \par
The complexity of our new minimax problem is a large consideration for the applicability of our results to large scale real-world situations. 
To address this, we use second order information to solve the inner optimization problem efficiently in terms of computational time. Solving this problem well means satisfying first order optimality conditions. Following~\cite{Beck2014}, we first negate the loss to produce an equivalent minimization problem and set up a Lagrangian.  \begin{align}\label{eq:LagrangianTrue}
        \Lcal(\bfdelta_{\bfx},\lambda) = -L(f_{\bftheta}(\bfx + \bfdelta_{\bfx}),\bfy) + \lambda (\tfrac{1}{2}\|\bfdelta_{\bfx}\|_2^2 - \tfrac{1}{2}r^2)
    \end{align}
Here $\lambda$ is a Lagrangian multiplier and we use an equivalent version of the constraint $\frac{1}{2}\|\bfdelta_{\bfx}\|_2^2 \le \frac{1}{2}r^2$ that we can differentiate more easily. 
The perturbation $\bfdelta_{\bfx}$ that maximizes the inner optimization problem of~\eqref{eq:bg-robustOptimization} necessarily satisfies the Karush-Kuhn-Tucker (KKT) conditions below~\cite{NocedalWright2006}.
    \begin{subequations}\label{eq:kkt}
    \begin{align}
        \nabla_{\bfdelta_{\bfx}} \Lcal(\bfdelta_{\bfx}, \lambda) 
            = -\nabla_{\bfdelta_{\bfx}}L(f_{\bftheta}(\bfx + \bfdelta_{\bfx}),\bfy) + \lambda \bfdelta_{\bfx} &= \bf0 && \text{(stationarity)} \label{eq:kkt-stationarity} \\
        \|\bfdelta_{\bfx}\|_2 &\le r  
         && \text{(primal feasibility)}\label{eq:kkt-primal}\\ 
        \lambda &\ge 0 && \text{(dual feasibility)} \label{eq:kkt-dual}\\ 
        \lambda(\|\bfdelta_{\bfx}\|_2 - r) &= 0   && \text{(complementary slackness)}\label{eq:kkt-slackness} 
    \end{align}
    \end{subequations}
Satisfying the KKT conditions ensures that gradients of the outer optimization problem are accurate; in particular, for each training sample, we have
    \begin{align} \label{eq:bg-gradexpansion}
        \nabla_{\bftheta} L(f_{\bftheta}(\bfx + \bfdelta_{\bfx}(\bftheta)),\bfy) &= 
        \left[\nabla_{\bftheta'} f_{\bftheta'}(\bfx + \bfdelta_{\bfx}(\bftheta))\nabla_{f}L(f_{\bftheta'}(\bfx + \bfdelta_{\bfx}(\bftheta)),\bfy)\right]_{\bftheta' =\bftheta} \\
        &\quad + 
        \left[\nabla_{\bftheta'} \bfdelta_{\bfx}(\bftheta')\nabla_{\bfdelta_{\bfx}}L(f_{\bftheta}(\bfx + \bfdelta_{\bfx}(\bftheta')),\bfy)\right]_{\bftheta' = \bftheta} \notag
    \end{align}
The first term in~\cref{eq:bg-gradexpansion} is the traditional gradient that we want to preserve.  
The second term comes from considering the perturbation as a function of the network weights, $\bfdelta_{\bfx}(\bftheta)$. From the stationarity condition~\eqref{eq:kkt-stationarity}, we get that $\nabla_{\bfdelta_{\bfx}}L(f_{\bftheta}(\bfx + \bfdelta_{\bfx}),\bfy)$ is parallel to $\bfdelta_{\bfx}$ if the perturbation is a maximizer. 
If the constraint is inactive ($\lambda = 0$), then $\nabla_{\bfdelta_{\bfx}}L(f_{\bftheta}(\bfx + \bfdelta_{\bfx}),\bfy) = \bf0$ and the second term is zero. 
If the constraint is active ($\lambda > 0$), then
from primal feasibility~\eqref{eq:kkt-primal} we know that the perturbation must satisfy the constraint even when undergoing changes incurred from $[\nabla_{\bftheta'} \bfdelta_{\bfx}(\bftheta')]_{\bftheta' = \bftheta}$.
With a sufficiently small perturbation of the weights $\bftheta$, the change in perturbation will follow the boundary of the constraint, nearly orthogonal to the direction of the gradient $\nabla_{\bfdelta_{\bfx}}L(f_{\bftheta}(\bfx + \bfdelta_{\bfx}),\bfy)$. 
This again makes the second term zero. Thus, if we solve the inner optimization problem well and thereby satisfy the KKT conditions, we can ignore the contribution of the second term. 

    \subsection{Fairness}
    \label{sec:fairness}
    We use three different fairness metrics defined in \cite{barocas-hardt-narayanan} in our experiments. All of these fairness metrics pertain to fairness of a classifier with respect to a sensitive attribute, in terms of true labels against the classifier's predictions. In all of our experiments, the sensitive attribute $s$, true label $Y$, and classifier prediction $\hat{Y}$ are all binary. For convenience, in defining the fairness metrics, we treat $Y$ as a random variable representing an object's true label and $\hat{Y}$ as a random variable representing its prediction.

\subsubsection{Independence}
For a classifier to satisfy independence its prediction $\hat{Y}$ must be uncorrelated with the sensitive attribute $s$. This requires an equal rate of positive classifications across all sensitive groups.
\begin{equation}
    P(\hat{Y} = 1|s = 0) = P(\hat{Y} = 1|s = 1) = P(\hat{Y} = 1)
    \label{eq:bg-indepedence}
\end{equation}
For instance, if the classifier was being used to recommend hiring decisions (so $\hat{Y}=1$ means a candidate should be hired, and $\hat{Y}=0$ means a candidate should not), satisfying independence would mean that if the classifier hires $20\%$ of applicants in class $s=1$, then it also hires $20\%$ of applicants in class $s=0$. 

\subsubsection{Separation}
Separation is similar to independence; for separation to be satisfied $\hat{Y}$ must be conditionally independent of $s$ given the value of $Y$.
\begin{align}
    \label{eq:bg-separation}
    P(\hat{Y} = 1|Y = 1, s = 0) = P(\hat{Y} = 1|Y = 1, s = 1) \\
    P(\hat{Y} = 1|Y = 0, s = 0) = P(\hat{Y} = 1|Y = 0, s = 1) \notag
\end{align}
Separation enforces equality of true and false positive rates. If again $\hat{Y}$ determines hiring recommendations, then $Y$ might indicate an individual’s true qualifications. Separation requires that individuals with similar qualifications have an equal chance of being hired, regardless of sensitive attribute. 

\subsubsection{Sufficiency}
Sufficiency enforces the conditional independence of $Y$ and $s$ given $\hat{Y}$.
\begin{align}
    P(Y = 1|\hat{Y} = 1, s = 0) = P(Y = 1|\hat{Y} = 1, s = 1)
    \label{eq:bg-sufficiency} \\
    P(Y = 1|\hat{Y} = 0, s = 0) = P(Y = 1|\hat{Y} = 0, s = 1) \notag
\end{align}
Sufficiency requires that the rates of individuals with the same predicted label also having the same true label is equal across different sensitive groups. If sufficiency is satisfied, then an individual from one group who is hired by the classifier is as likely to be truly qualified as a hired individual from another group.  

\section{Our Approach} Next we introduce our proposed second order method, and discuss its implementation. Our approach relies on approximation, so an analysis of the error produced by this approximation follows in \cref{sec:analysis}. Then in \cref{sec:Other Methods}, we introduce alternate methods of solving the robust optimization problem and their implementations to test against our proposed approach.
    \label{sec:our_appraoch}
    
    \subsection{Trust Region Subproblem (TRS)}
    \label{sec:trust_region_subproblem}
    \begin{algorithm}[t]
\caption{Trust Region Subproblem}\label{alg:TRS}
\begin{algorithmic}[1]
\Require Loss function $f: \mathbb{R}^n \to \mathbb{R}$, point $\bm{x} \in \mathbb{R}^n$, trust region radius $\delta$
\Ensure Approximate solution $\bm{x}^*$ and step $\bm{s}$

\State Evaluate $f(\bm{x})$, $\nabla f(\bm{x})$, and $\nabla^2 f(\bm{x})$
\State Compute Newton step $\bm{s} = -\nabla^2 f(\bm{x})^{-1} \nabla f(\bm{x})$

\If{$\delta$ is not specified}
    \State Set $\delta$ to $\|\bm{s}\|$
\EndIf

\If{per sample}
    \For{each sample $i$}
        \If{$\|\bm{s}[i]\| > \delta$}
            \State Set $\lambda_{\text{low}} = 0$, compute $\lambda_{\text{high}}$
            \State solve for $\lambda$ using bisection method on the problem $\|\bm{s}_\lambda\| - \delta = 0$
            \State Update step $\bm{s}[i]$ based on the solution
        \EndIf
    \EndFor
\EndIf
\State Update $\bm{x}^* = \bm{x} + \bm{s}$

\Return $\bm{x}^*$, $\bm{s}$
\end{algorithmic}
\end{algorithm}
Our main algorithm (\cref{alg:TRS}) solves an approximation of the inner optimization problem \eqref{eq:bg-innerOptimization} using second order information.
For each training sample $(\bfx, \bfy)\in \Tcal$, we fix $\bftheta$ and expand the loss function using a quadratic Taylor series approximation about $\bfx$ in the direction of $\bfdelta_{\bfx}$.
\begin{equation}
         \min_{\|\bfdelta_{\bfx}\|_2 \le r} -L(f_{\bftheta}(\bfx),\bfy) -(\nabla_{\bfx} L(f_\bftheta(\bfx),\bfy))^T \bfdelta_{\bfx} - \tfrac{1}{2}\bfdelta_{\bfx}^T \nabla^2_{\bfx} L(f_{\bftheta}(\bfx),\bfy) \hspace{2pt} \bfdelta_{\bfx}
         \label{eq:ourapp-TRS-quadratic approx}
\end{equation}
To fit our constraint, we construct a Lagrangian term by squaring our initial constraint and scaling the Lagrange multiplier by one-half. This gives us a function that depends on $\bfdelta_{\bfx}$ and $\lambda$. 
\begin{equation}
    \tilde{\Lcal}(\bfdelta_{\bfx}, \lambda) = -L(f_{\bftheta}(\bfx),\bfy) - (\nabla_{\bfx} L(f_\bftheta(\bfx),\bfy))^T \bfdelta_{\bfx} - \frac{1}{2}\bfdelta_{\bfx}^T \nabla^2_{\bfx} L(f_{\bftheta}(\bfx),\bfy) \hspace{2pt} \bfdelta_{\bfx} + \dfrac{\lambda}{2}(||\bfdelta_{\bfx}||^2 - r^2)
    \label{eq:ourapp-TRS-lagrangian}
\end{equation}
We approximate the optimal $\bfdelta_{\bfx}^*$ to the inner optimization problem as the optimal $\bfdelta_{\bfx}$ solution to the quadratic problem \eqref{eq:ourapp-TRS-quadratic approx}. 
The KKT conditions are the same as~\eqref{eq:kkt} except for the stationarity condition.
    \begin{equation}\label{eq:ourapp-quadApprox-stationarity}
         - \nabla_{\bfx} L(f_\bftheta(\bfx),\bfy) -\nabla^2_{\bfx} L(f_{\bftheta}(\bfx),\bfy) \hspace{2pt} \bfdelta_{\bfx} + \lambda\bfdelta_{\bfx} = \bf0 \hspace{4em} \text{(stationarity)}
    \end{equation}
This gives us an explicit relation of $\bfdelta_x$ to $\lambda$.
\begin{equation}
    \bfdelta_{\bfx}(\lambda) = -(\nabla^2_{\bfx} L(f_{\bftheta}(\bfx),\bfy) -\lambda I)^{-1}\nabla_{\bfx} L(f_\bftheta(\bfx),\bfy)
    \label{eq:ourapp-stationarity-delta_x}
\end{equation}
There are two cases to \eqref{eq:ourapp-stationarity-delta_x}. If $\lambda = 0$, then the optimal $\bfdelta_{\bfx}$ for \eqref{eq:ourapp-TRS-quadratic approx} can be found by solving a system of linear equations involving the gradient and the Hessian. Alternatively, if $\lambda \ne 0$, then complementary slackness enforces $\|\bfdelta_{\bfx}\|_2 = r$, so we need to find $\lambda$ such that $\|\bfdelta_{\bfx}(\lambda)\|_2 = r$.

\subsubsection{The Bisection Method Bracket} 
To find a value for $\lambda$ such that $\|\bfdelta_{\bfx}(\lambda)\|_2 = r$, we build a univariate function $g(\lambda):= \|\bfdelta_{\bfx}(\lambda)\|_2 - r$ and find a root of this function. Applying some linear algebra to \eqref{eq:ourapp-stationarity-delta_x}, one can show that
\begin{subequations}
\begin{align}
    g(\lambda) &= \|\bfdelta_{\bfx}(\lambda)\|_2 - r\\
    &=\|-(\nabla^2_{\bfx} L(f_{\bftheta}(\bfx),\bfy) -\lambda I)^{-1}\nabla_{\bfx} L(f_\bftheta(\bfx),\bfy)\|_2 - r\\
    &=\|(QDQ^\top -\lambda I)^{-1}\nabla_{\bfx} L(f_\bftheta(\bfx),\bfy)\|_2 - r\\
    &=\|Q(D -\lambda I)^{-1}Q^\top\nabla_{\bfx} L(f_\bftheta(\bfx),\bfy)\|_2 - r\\
    &=\|(D -\lambda I)^{-1}Q^\top\nabla_{\bfx} L(f_\bftheta(\bfx),\bfy)\|_2 - r
\end{align}
\end{subequations}
where $\nabla_{\bfx}^2 L(f_{\bftheta}(\bfx), \bfy) = Q D Q^T$ is the eigendecomposition of the Hessian. Because the Hessian is symmetric, by the Spectral Theorem,  we know $Q$ is orthogonal and $D$ is diagonal and real-valued. 

We use the bisection method \cite{kaw} to find a root of $g(\lambda) = 0$. In order to do so, we need to establish a bracket $[\lambda_{\rm low}, \lambda_{\rm high}]$ such that $g$ has different signs at the endpoints; that is, $g(\lambda_{\rm low}) g(\lambda_{\rm high}) < 0$.
If $\lambda = 0$, then constraint is satisfied. 
Thus, $g(0) \ge 0$ and, in practice, positive, so $\lambda_{\rm low} = 0$ is a good candidate. 
To find the upper bound, we first bound the norm
    \begin{align}
        \|(D -\lambda I)^{-1}Q^\top\nabla_{\bfx} L(f_\bftheta(\bfx),\bfy)\|_2
        &\le \|(D -\lambda I)^{-1}\|_2 \|\nabla_{\bfx} L(f_\bftheta(\bfx),\bfy)\|_2\\
        &=\sqrt{\sum_{i=1}^{n_{\rm in}} \frac{1}{(d_i - \lambda)^2}} \|\nabla_{\bfx} L(f_\bftheta(\bfx),\bfy)\|_2
    \end{align}
Following from~\cite{NocedalWright2006}, as 
$\lambda \to d_{\rm max}^+$, the upper bound approaches $+\infty$ and as $\lambda \to \infty$, the upper bound approaches $0$. 
This guarantees that there is some $\lambda \in (d_{\rm max}, \infty)$ such that the upper bound is less than $r$. 
If we let 
    \begin{align}
        \lambda_{\rm high} = |d_{\rm max}| + \frac{\sqrt{n}\|\nabla_{\bfx} L(f_\bftheta(\bfx),\bfy)\|_2}{r}
    \end{align}
then $(d_i - \lambda_{\rm high})^2 \ge \dfrac{n\|\nabla_{\bfx} L(f_\bftheta(\bfx),\bfy)\|_2^2}{r^2}$ for $i=1,\dots,n_{\rm in}$. 
Substituting into the upper bound, we get
    \begin{align}
        \sqrt{\sum_{i=1}^{n_{\rm in}} \frac{1}{(d_i - \lambda)^2}} \|\nabla_{\bfx} L(f_\bftheta(\bfx),\bfy)\|_2
        &\le \frac{\sqrt{n}r}{\sqrt{n}\|\nabla_{\bfx} L(f_\bftheta(\bfx),\bfy)\|_2} \|\nabla_{\bfx} L(f_\bftheta(\bfx),\bfy)\|_2 = r.
    \end{align}
Thus, we have found a bracket for the bisection method.

    \subsection{Algorithm Analysis}
    \label{sec:analysis}
    
When solving the inner optimization problem using a second order approximation, we would like to know how well this approximation actually solves this problem. For specific classes of models, loss functions, and activation functions, we can confine the error explicitly to depend on high orders of the perturbation $\bfdelta_{\bfx}$ and loss function derivatives.

\subsubsection{Affine model} 
\begin{figure}[t]
    \centering\includegraphics[height=2in]{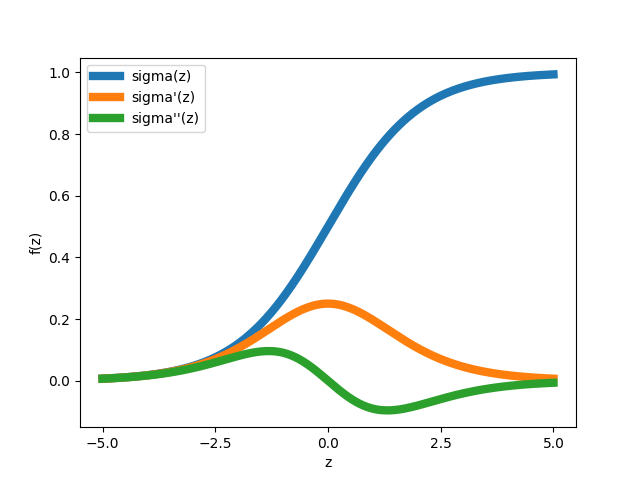}
    \caption{Flattening behavior of sigmoidal function, $\sigma$, derivatives.}
    \label{fig:ourapp-sigmoid-flat}
\end{figure}

An affine model $f_{\bfw,b}(\bfx) = \bfw^\top\bfx + b$ with weight vector $\bfw$ and a scalar bias $b$ combined with a logistic regression loss function $L$ and a sigmoid activation function $\sigma$ is convex with respect to inputs. 
The loss is given explicitly as
\begin{equation}
    \label{analysis-affine-sigmoid-loss}
    L(f_{\bfw,b}(\bfx),y) = -y\ln[\sigma(\bfw^\top\bfx + b)]-(1-y)\ln[1-\sigma(\bfw^\top\bfx + b)] 
\end{equation}
with $\sigma(z) = (1+e^{-z})^{-1}$. 
Introducing the perturbation $\bfdelta_{\bfx}$ results in a specific case of the inner optimization problem in~\eqref{eq:bg-innerOptimization}. To solve this optimization problem without approximation, we introduce as before a Lagrange multiplier $\lambda$ with the constraint $\lVert \bfdelta_{\bfx} \rVert^2 -r^2 \leq 0$.
\begin{equation}
\label{eq:analysis-loss}
\begin{split}
    \mathcal{L}_{\rm aff}(\bfdelta_{\bfx},\lambda) = -y\ln(\sigma(\bfw^\top(\bfx + \bfdelta_{\bfx}) + b))-(1-y)\ln(1-\sigma(\bfw^\top(\bfx + \bfdelta_{\bfx})+ b)) \\+ \frac{1}{2}\lambda(\| \bfdelta_{\bfx} \|_2^2 - r^2)
\end{split}
\end{equation}
The first order optimality conditions for \eqref{eq:analysis-loss} tell us that at the optimal $\bfdelta_{\bfx}$,
\begin{equation}
    \nabla_{\bfdelta_{\bfx}}\mathcal{L}_{\rm aff}(\bfdelta_{\bfx},\lambda) = 
    (-y+\sigma(\bfw^\top(\bfx + \bfdelta_{\bfx})+b))\bfw + \lambda\bfdelta_{\bfx} = {\bf 0}.
    \label{eq:analysis-actual}
\end{equation}
To compare the exact solution from equation \eqref{eq:analysis-actual} to the approximation made when solving using the trust region method of section \ref{sec:trust_region_subproblem}, we find the second order approximation of the loss function $L(f_{\bfw,b}(\bfx), y)$ by a Taylor expansion in $\bfx$ in the direction of $\bfdelta_{\bfx}$. 
\begin{equation}
    L(f_{\bfw,b}(\bfx+\bfdelta_{\bfx}),y) \approx L(f_{\bfw,b}(\bfx),y) + \nabla_\bfx^\top L(f_{\bfw,b}(\bfx),y)\bfdelta_{\bfx} + \frac{1}{2}\bfdelta_{\bfx}^\top \nabla_\bfx^2 L(f_{\bfw,b}(\bfx),y) \bfdelta_{\bfx} 
\end{equation}
where the gradient and Hessian are
\begin{equation}
    \begin{split}
        \nabla_\bfx L(f_{\bfw,b}(\bfx),y) &= (-y + \sigma(\bfw^\top\bfx + b))\bfw \\
        \nabla_\bfx^2 L(f_{\bfw,b}(\bfx), y) &= \bfw\sigma'(\bfw^\top\bfx + b)\bfw^\top.
    \end{split}
\end{equation}
The associated Lagrangian is
\begin{equation}
    \tilde{\mathcal{L}}_{\rm aff}(\bfdelta_{\bfx},\lambda) = L(f_{\bfw,b}(\bfx),y) + \nabla_\bfx^\top L(f_{\bfw,b}(\bfx),y)\bfdelta_{\bfx} + \frac{1}{2}\bfdelta_{\bfx}^\top \nabla_\bfx^2 L(f_{\bfw,b}(\bfx),y) \bfdelta_{\bfx} + \frac{1}{2}\lambda(\lVert \bfdelta_{\bfx} \rVert^2 - r^2).
\end{equation}
As before, take the gradient with respect to $\bfdelta_{\bfx}$ and set it equal to zero to solve using first-order optimization conditions.
\begin{equation}
    \label{eq:analysis-approx}
    \begin{split}
        \nabla_{\bfdelta_{\bfx}}\mathcal{L}(\bfdelta_{\bfx},\lambda) 
        &= (-y + \sigma(\bfw^\top\bfx + b))\bfw + \bfw\sigma'(\bfw^\top\bfx + b)\bfw^\top\bfdelta_{\bfx} + \lambda\bfdelta_{\bfx} = \bf0
    \end{split}
\end{equation}
Comparing \eqref{eq:analysis-actual} (\textcolor{cyan}{LHS}) and \eqref{eq:analysis-approx} (\textcolor{orange}{RHS}), the discrepancy between the exact solution and the approximation is restricted to
\begin{equation}
   \textcolor{cyan}{\sigma(\bfw^\top(\bfx + \bfdelta_{\bfx})+b)} \neq \textcolor{orange}{\sigma(\bfw^\top\bfx + b) + \sigma'(\bfw^\top\bfx + b)\bfw^\top\bfdelta_{\bfx}}.
   \label{eq:analysis-comp}
\end{equation}
Now, appling a Taylor expansion centered at $\bfx$ in the ball $\bfx + \bfdelta_{\bfx}$ to the \textcolor{cyan}{LHS}, we obtain:
\begin{equation}
    \label{eq:analysis-comp-expansion}
   \textcolor{cyan}{\sigma(\bfw^\top(\bfx + \bfdelta_{\bfx})+b)} \approx \textcolor{orange}{\sigma(\bfw^\top\bfx + b) + \sigma'(\bfw^\top\bfx + b)\bfw^\top\bfdelta_{\bfx}} + \frac{1}{2}\bfdelta_{\bfx}^\top \bfw\sigma''(\bfw^\top\bfx + b)\bfw^\top\bfdelta_{\bfx}
\end{equation}
We have recovered the \textcolor{orange}{RHS} of \eqref{eq:analysis-comp}, so the error is due to the truncation of the second-order and higher terms of \eqref{eq:analysis-comp-expansion}. In this case with a sigmoidal loss function, that means this error depends on the magnitude of $|\sigma''(z)|$ and the $\bfdelta_{\bfx}$ for which we solved. 
For any sigmoidal function, their structure gives first and second order derivatives of the classes shown in \cref{fig:ourapp-sigmoid-flat}. For our sigmoid function defined as $\sigma(z)=(1+e^{-z})^{-1}$ with $|\sigma''(z)| \le 0.1$, and in general for any choice of sigmoidal function this flattening of higher-order derivatives will be observed. By nature $\bfdelta_{\bfx}$ is bounded by the data since it defines the perturbation from a given point, and a perturbation larger than the size of the data space in any given dimension would be meaningless. For data in the form of continuous numerical values normalized to be between $0$ and $1$, as in our case, each component of $\bfdelta_\bfx$ must be less than $1$. In practice, $\bfdelta_\bfx$ tends to be much smaller than that. This bound means higher order terms are generally quite small, and for this combination of loss function, activation function, model, and radii on the order of $10^{-1}$, the approximation error is on the order of $\|\bfdelta_{\bfx}\|^2 |\sigma''(z)| \approx 10^{-3}$.

    \subsection{Other Methods}
    \label{sec:Other Methods}
    We compare our trust region subproblem (TRS) algorithm to random perturbation to examine whether or not it is important to solve the inner optimization problem well.
We also use random perturbation, along with a projected gradient descent (PGD) method, to verify that our second order TRS approach has greater computational efficiency than only using lower order information.

\begin{subsubsection}{Random Perturbation} 
For each data point, we sample the perturbation $\boldsymbol{\delta}_x$ randomly from a multivariate standard normal distribution and rescale to the length of the trust region radius. 
This method acts as a control in our experiments to show the advantages of solving for an optimal perturbation. 
\end{subsubsection}

\begin{subsubsection}{Projected Gradient Descent (PGD)}
In order to test that our second order TRS approach is computationally faster than using purely first order information, we also implemented a version of gradient descent for our inner optimization problem. Since our inner problem has the constraint $\|\bfdelta_x\|_2 \le r$, we cannot use vanilla gradient descent, and instead use projected gradient descent (PGD) \cite{beck2017first, parikh2014proximal}. PGD operates similarly to standard gradient descent, but once it has found its optimal step it projects the step onto the constrained set before returning it. Mathematically, this looks like:
\begin{equation}
    \boldsymbol{\delta}_{\boldsymbol{x}}^{(k+1)} = P\left[\boldsymbol{\delta}_{\boldsymbol{x}}^{(k)}+\alpha^{(k)}\cdot \nabla_{\boldsymbol{x}} \hspace{1pt} L(f_{\boldsymbol{\theta}}(\boldsymbol{x} +\boldsymbol{\delta}_{\boldsymbol{x}}),\boldsymbol{y})\right]
\end{equation}
where $\boldsymbol{\delta}_{\boldsymbol{x}}^{(k)}$ is the $k$th iterate, $\alpha^{(k)}$ is the step size at the $k$th iteration, and P is the projection operator $P(\bfx) = \argmin_{\bfz: ||\bfz||_2 \le r} ||\bfx - \bfz||_2^2$. This projection operator turns out to be very simple, returning the inputted point if the point already satisfies the constraint, or scaling the point inward to the boundary of the constraint if it is outside it. In particular,
\begin{equation}
    P(x) = \min\left\{1, \dfrac{r}{||\bfx||_2}\right\} \bfx.
\end{equation}

Numerical experiments on how PGD performs with adversarial training \cite{xu2021robust, madry2018towards} have shown that PGD is a reliable method when it comes to solving robust optimization problems. 
\end{subsubsection}

\section{Numerical Results} Now we present results of our numerical experiments pertaining to fairness, accuracy, and computational time. Subsections \ref{sec:synthetic}, \ref{sec:adult}, and \ref{sec:lsat} discuss fairness and accuracy results on three datasets. For each dataset, we compute the fairness and accuracy results for nonrobust training, for robust training with various radii, and random perturbations. We measure the absolute difference across sensitive attributes for analysis of each fairness metric, and the closer this difference is to zero, the fairer the classifier is with respect to that metric. \Cref{sec:efficiency} presents our results on relative computational time across our various methods for solving the inner optimization problem.
\label{sec:numerical}

    \subsection{Synthetic Data (Unfair2D)}
    \label{sec:synthetic}
\begin{figure}[t]
    \centering
    \label{fig:num-synth-setup}
    \captionsetup[subfigure]{justification=raggedright}
    \begin{subfigure}[t]{0.49\linewidth}
        \includegraphics[width=\textwidth]{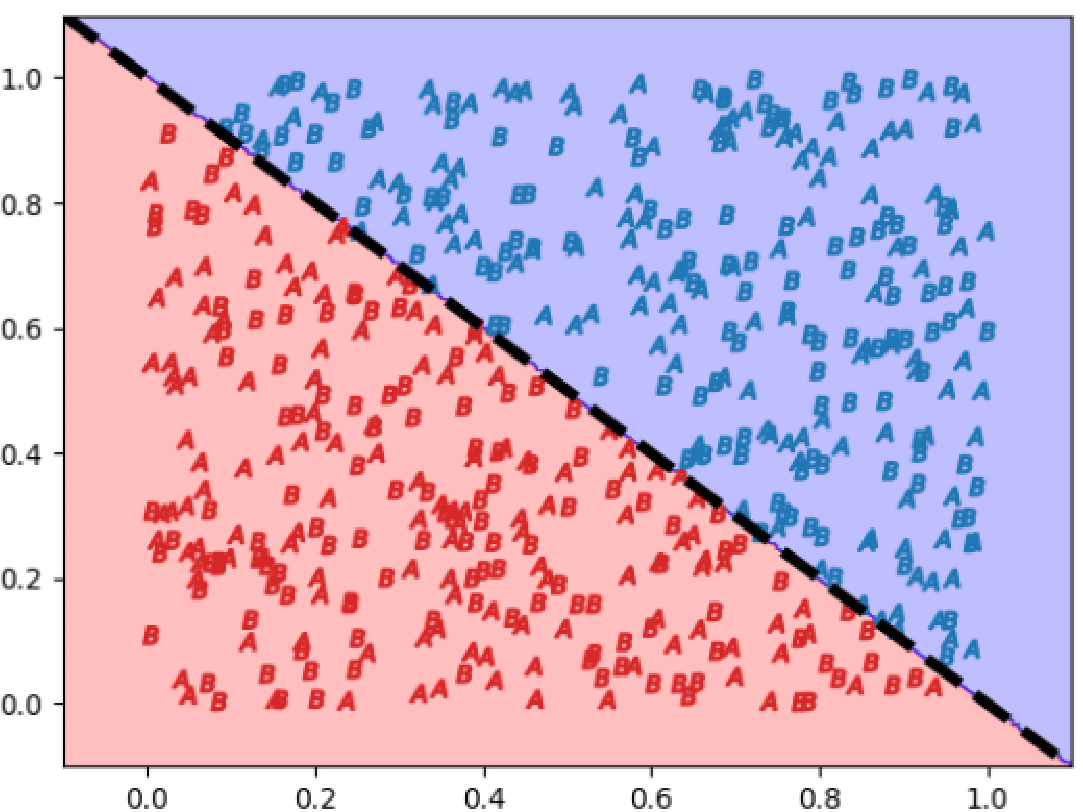}
        \caption{Pre-shift data}
        \label{fig:num-synth-setup-original}
    \end{subfigure}%
    ~
    \begin{subfigure}[t]{0.49\linewidth}
        \includegraphics[width=\columnwidth]{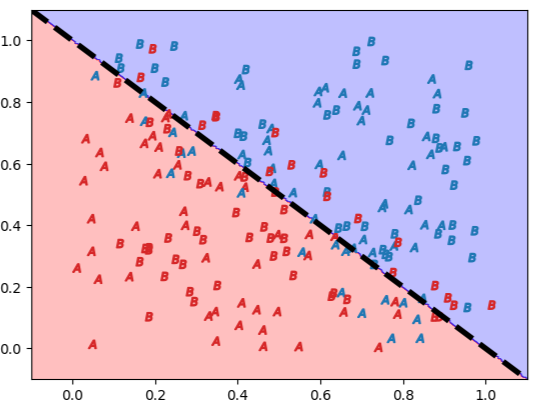}
        \caption{Sample of post-shift data}
        \label{fig:num-synth-setup-post}
    \end{subfigure}
    \caption{Points are colored based on their original location in the blue region ($Y = 1$) or red region ($Y = 0$). Post shift, note the unfair presence of red $B$s in the blue region and blue $A$s in the red region.}
\end{figure}

    The primary dataset we used for carrying out numerical experiments was a synthetic dataset. Individuals belonging to two different groups, labeled with respect to a sensitive attribute $A$ or $B$, are being hired on the basis of two numeric scores $x_1$ and $x_2$. Individuals have a binary label that is either ``should be hired'' or ``shouldn't be hired,'' and we train a linear classifier to decide whether or not to hire an individual. The data is initially fair (\cref{fig:num-synth-setup-original}), and we introduce unfair bias into the data by artificially raising the scores of all $B$s while lowering the scores of all $A$s  (\cref{fig:num-synth-setup-post}). In the real world, this could be a manifestation of structural unfairness in which $B$s are more likely to belong to a wealthy socioeconomic class, and thus can afford training that boosts their scores, whereas $A$s do not have this opportunity. In fact, $A$s may not only lack the advantage of $B$s, but also have an active disadvantage, such as an increased likelihood of needing to work longer hours, impeding time for study and test prep, lowering their scores.
\begin{table}[t]
    \centering
    \begin{tabular}{c c c c}
        \multicolumn{4}{c}{\includegraphics[width=0.4\linewidth]{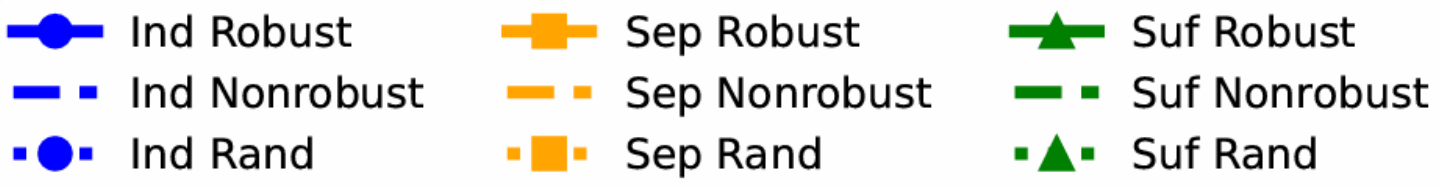}} \\
        & \small{$Y,\hat{Y}=0$} & \small{$Y,\hat{Y}=1$} & \small{Accuracy} \\
        \raisebox{3em}{\rotatebox{90}{Training set}} & \includegraphics[width=0.28\linewidth]{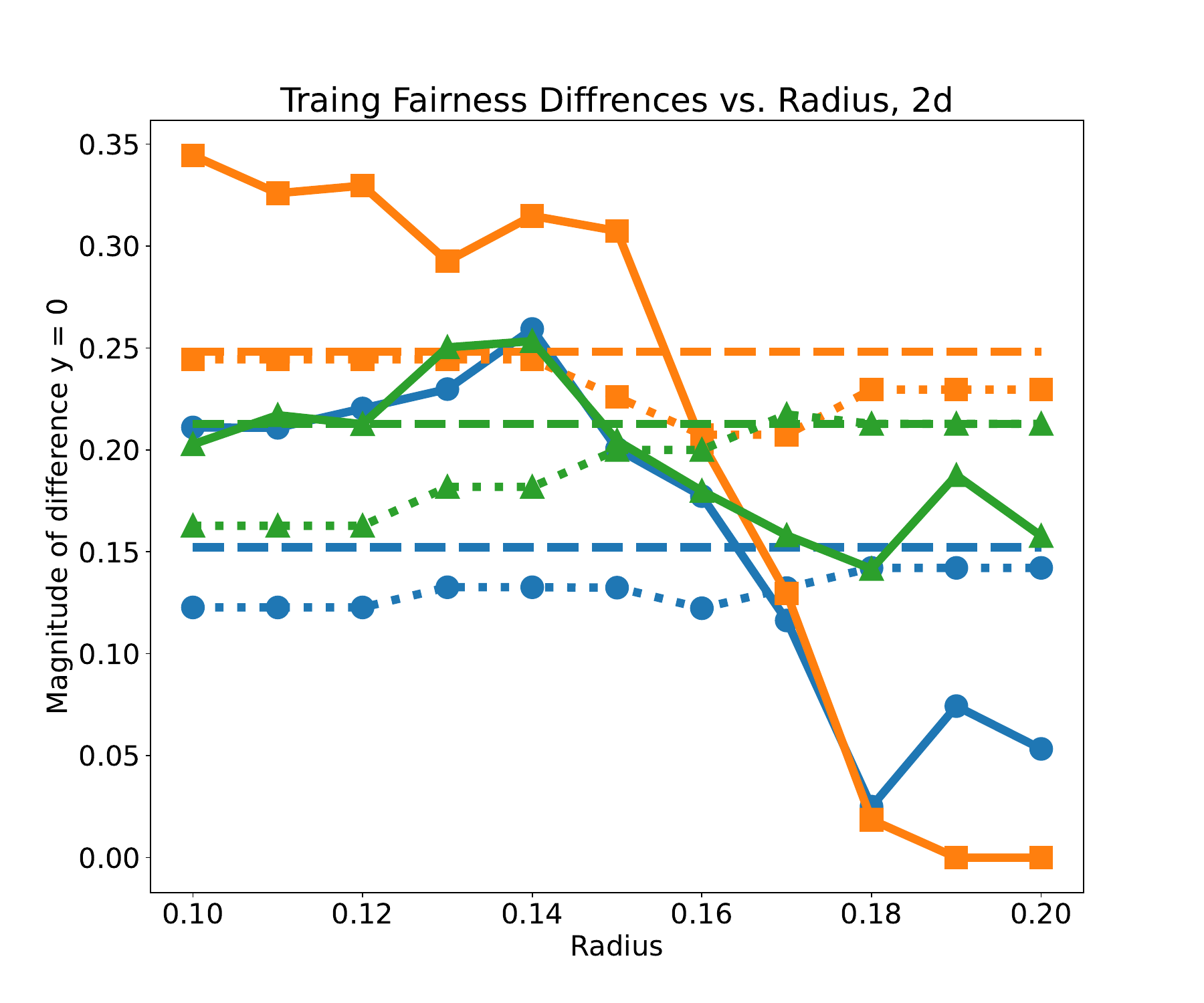} & \includegraphics[width=0.28\linewidth]{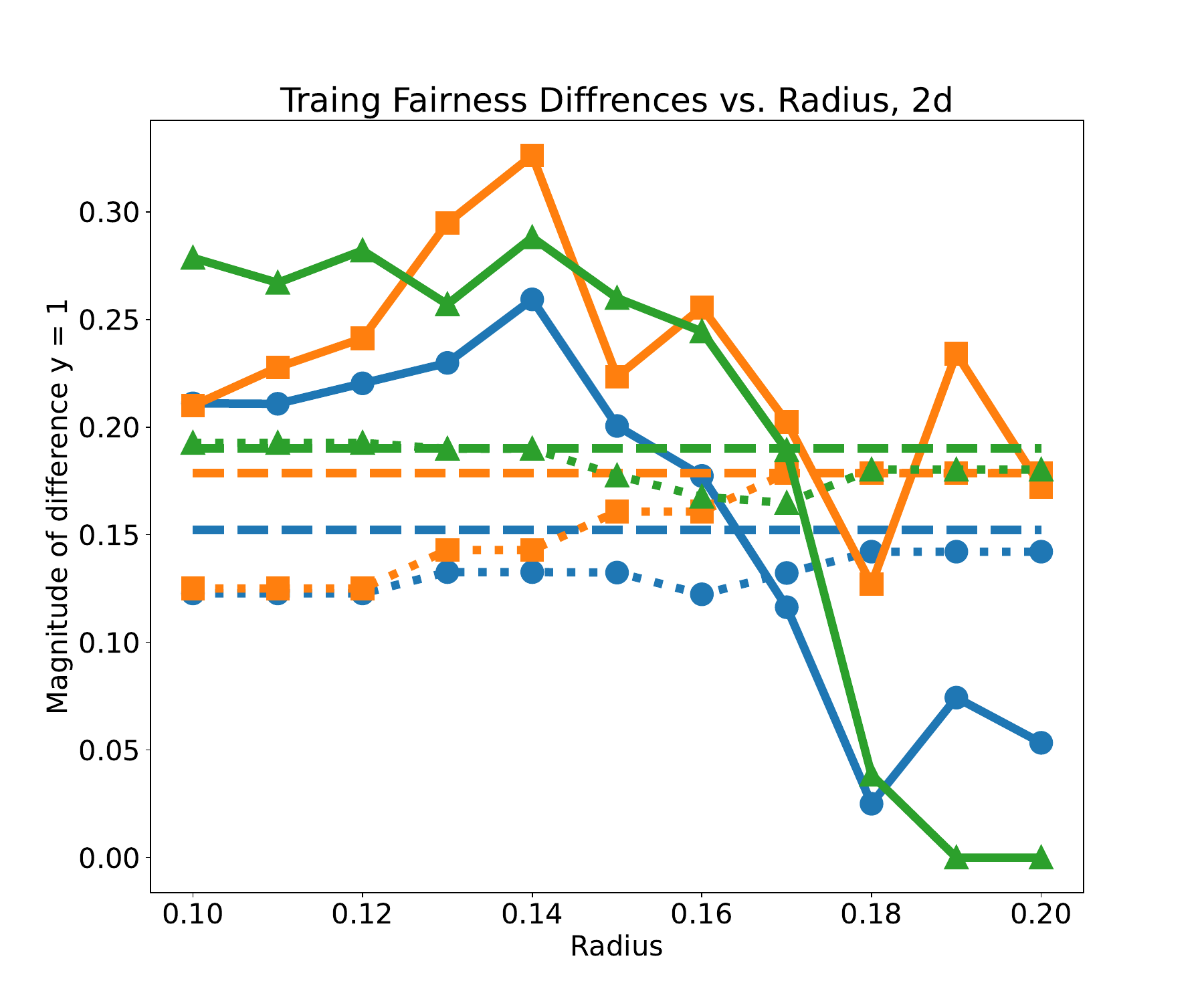}
        & \includegraphics[width=0.28\linewidth]{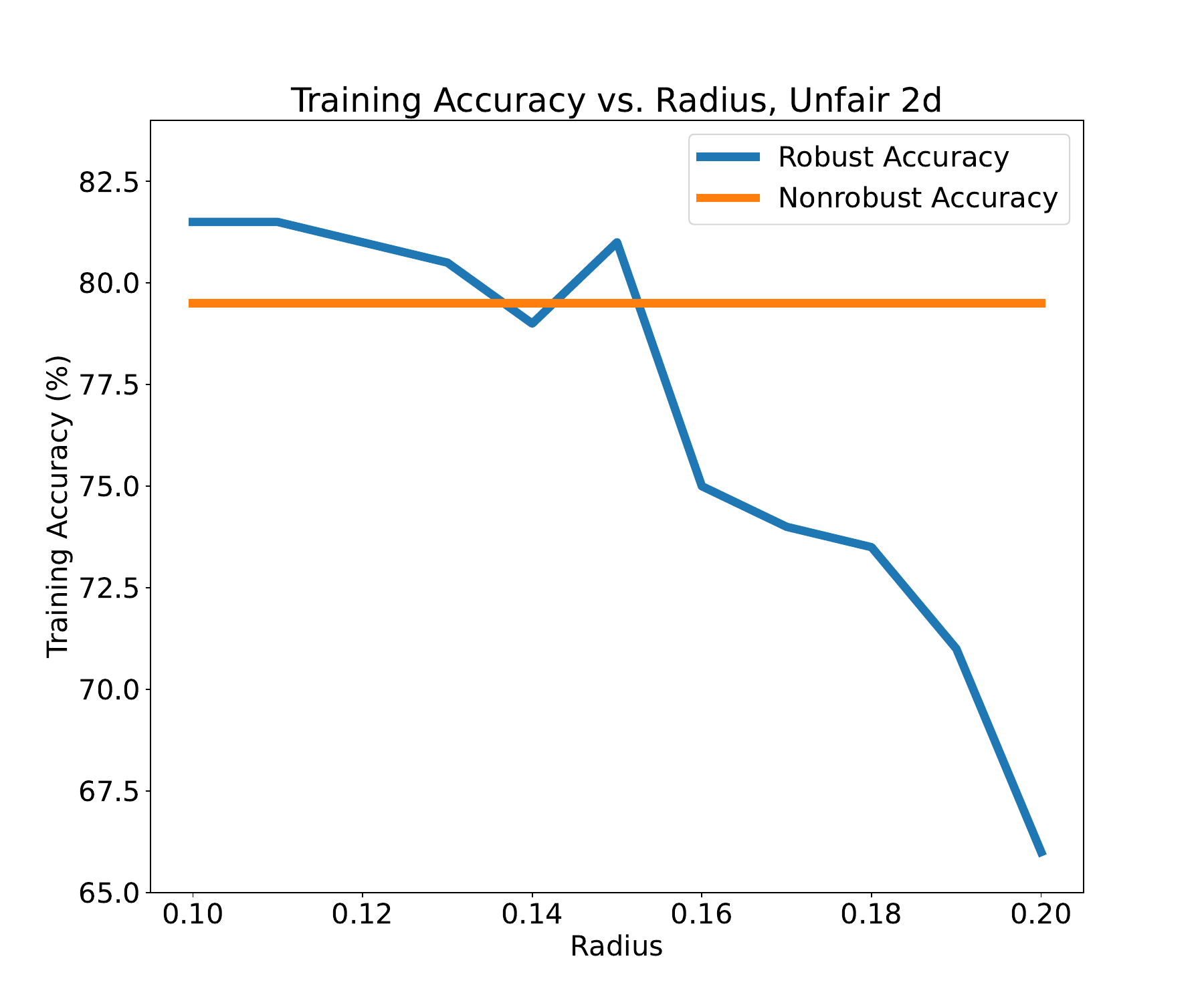} \\
        & (a) & (b) & (c) \\
        \raisebox{3em}{\rotatebox{90}{Test set}} & \includegraphics[width=0.28\linewidth]{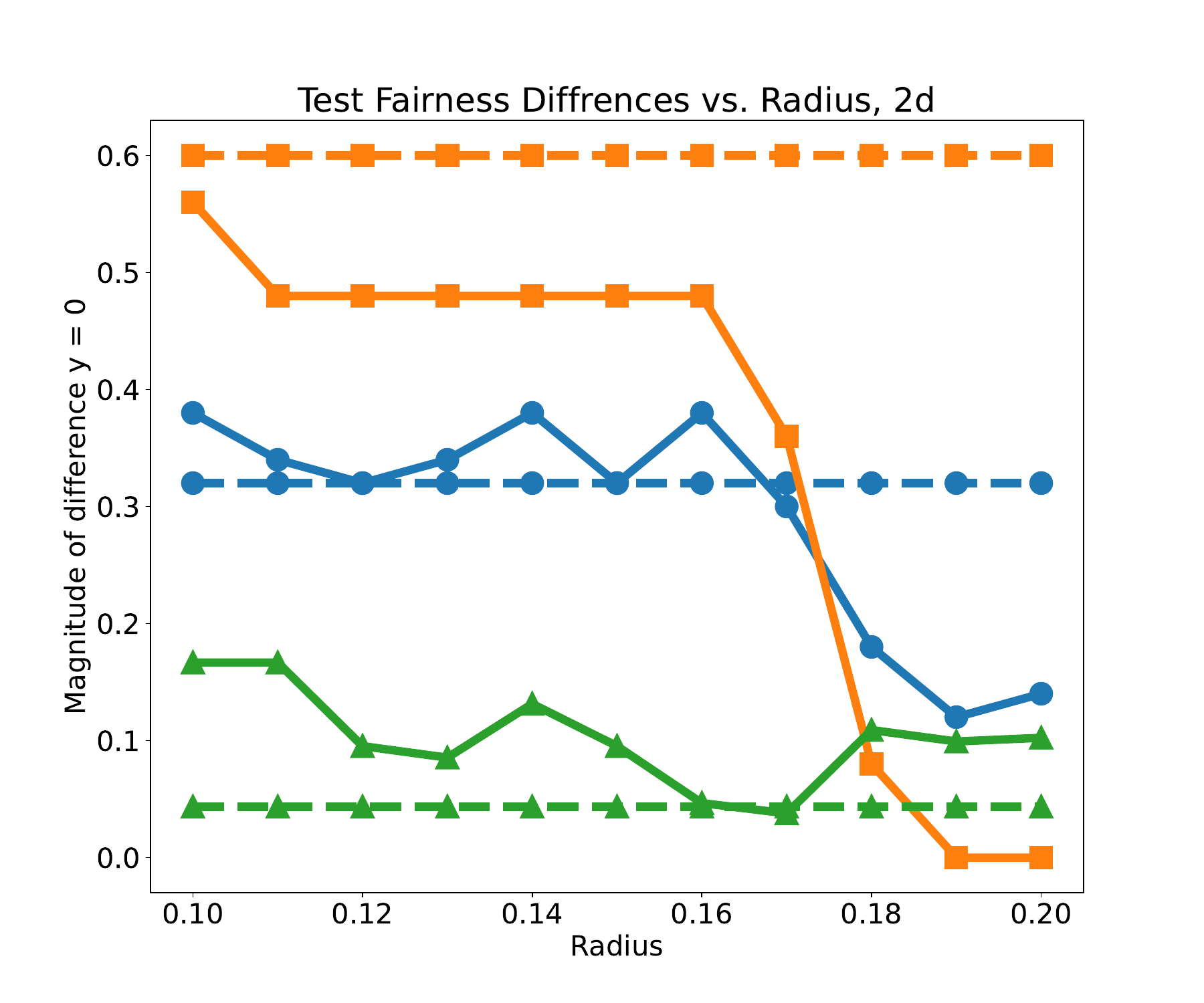} & \includegraphics[width=0.28\linewidth]{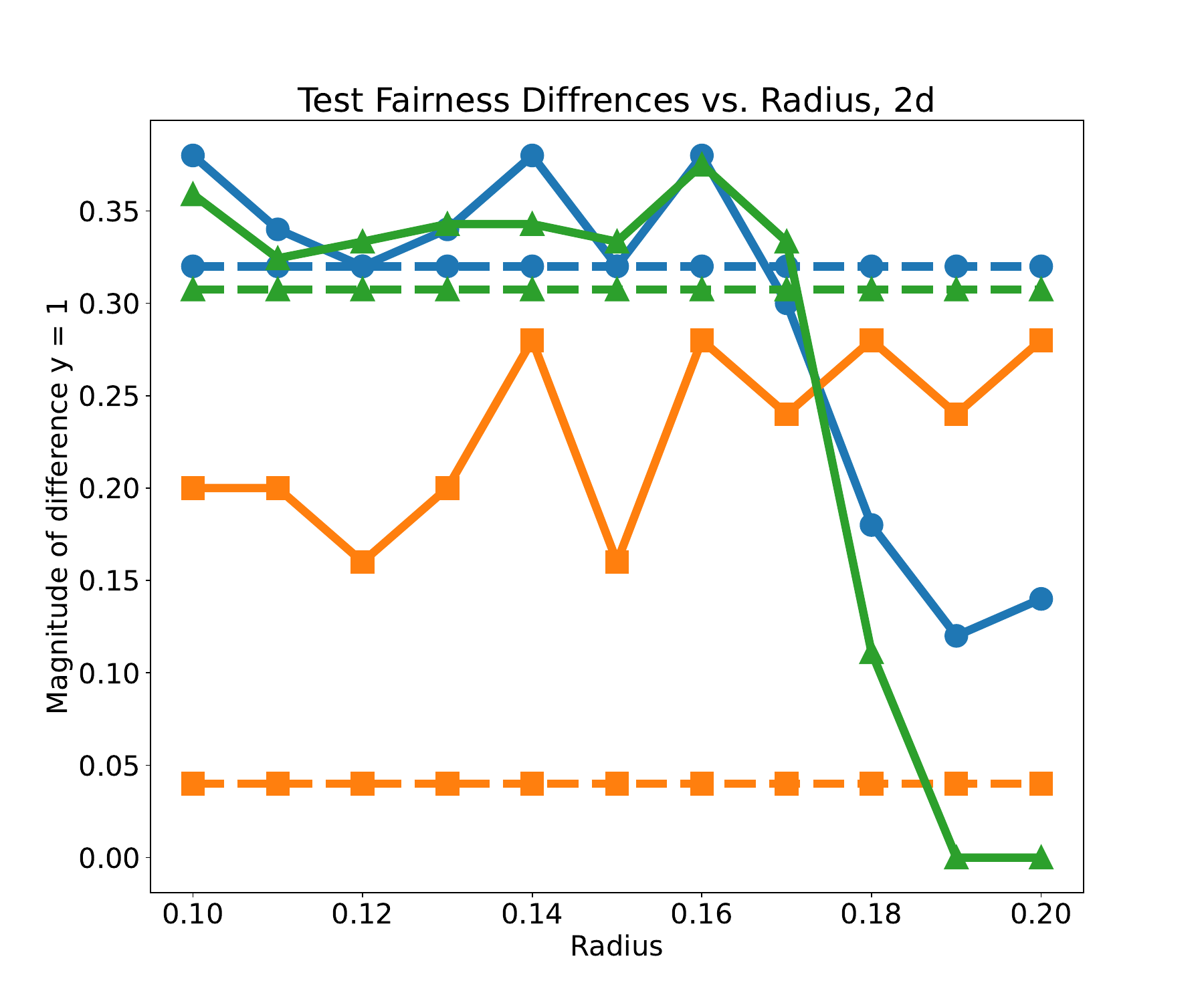} & \includegraphics[width=0.28\linewidth]{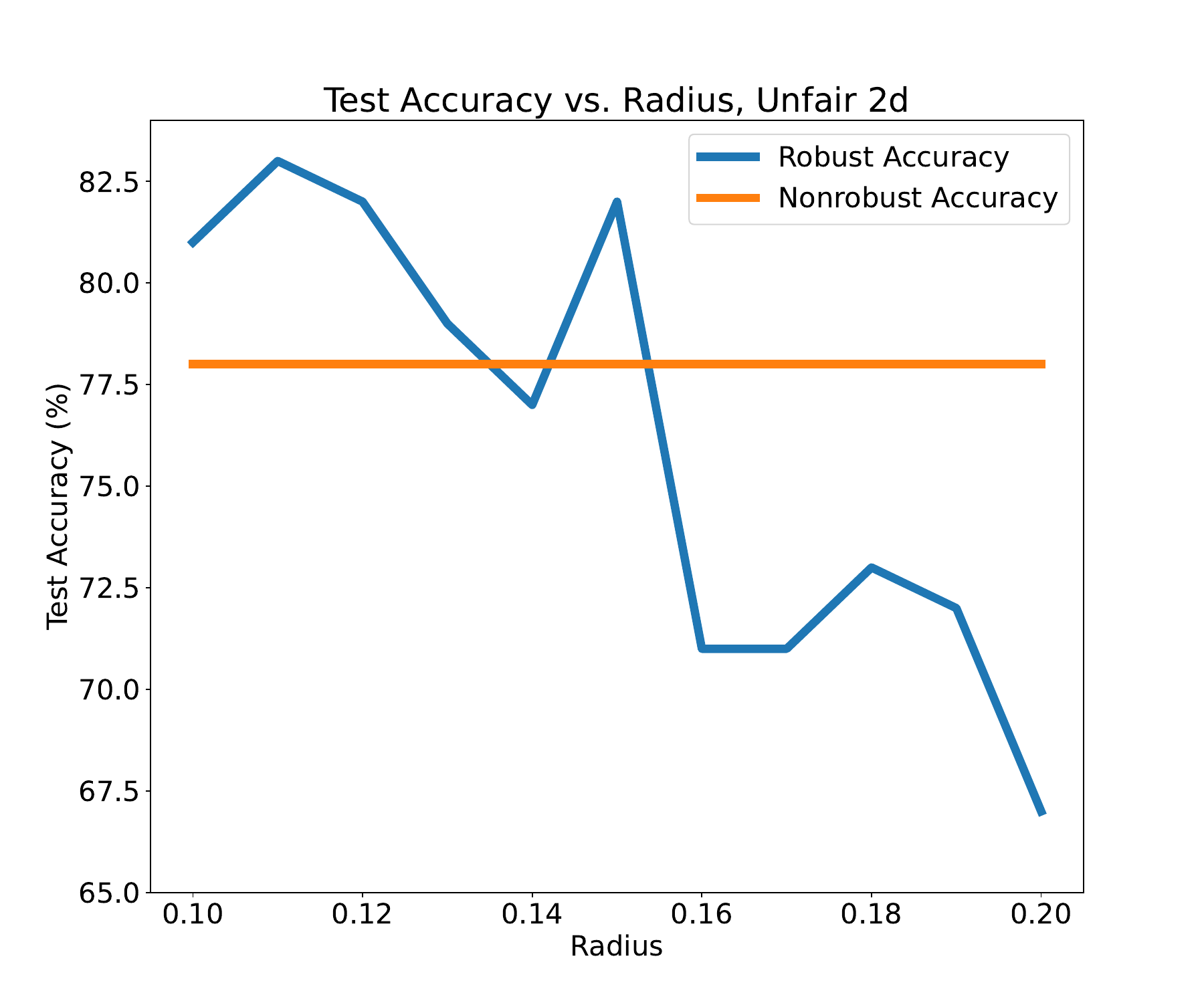} \\
        & (d) & (e) & (f)
    \end{tabular}
    \caption{Synthetic fairness and accuracy results. For fairness, values closer to zero are desirable.}
    \label{tab:num-syn-results}
\end{table}

\begin{figure}[t]
    \centering
    \begin{subfigure}[b]{0.5\textwidth}
        \centering
        \includegraphics[width=0.9\linewidth]{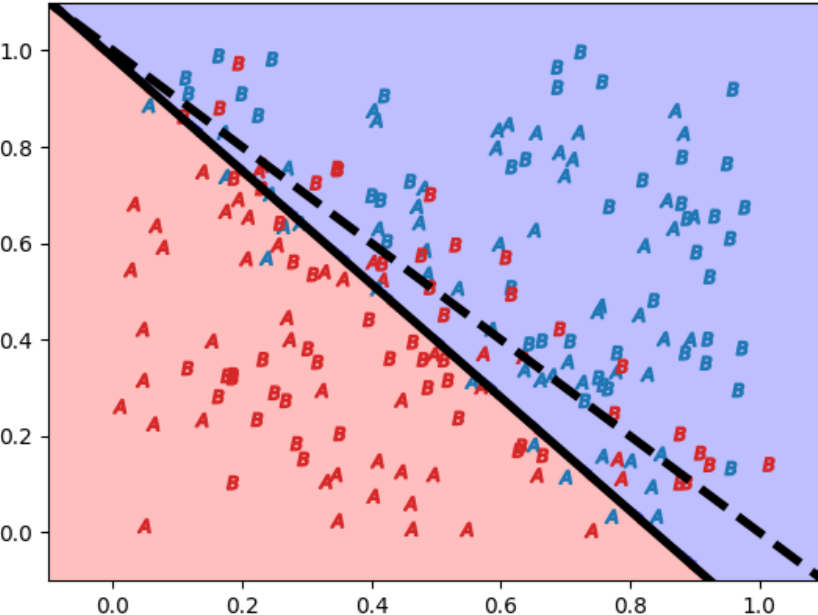}
        \caption{Nonrobust Classifier}
        \label{fig:num-syn-comp-Nonrobust}
        \begin{tabular}{|c|c|c|}
            \hline
            \textbf{Diff:} & $Y=0, |S1-S0|$ & $Y=1, |S1-S0|$ \\
            \hline
            \hline
            Ind. & 0.152 & 0.152 \\
            \hline 
            Sep. & 0.248 & 0.179 \\
            \hline
            Suff. & 0.213 & 0.190 \\
            \hline
        \end{tabular} \\
        \smallskip
        \textbf{Training Accuracy:} 79.5\% \\
        \textbf{Test Accuracy:} 78.0\%
    \end{subfigure}%
    \hfill
    \begin{subfigure}[b]{0.5\textwidth}
        \centering
        \includegraphics[width=0.9\linewidth]{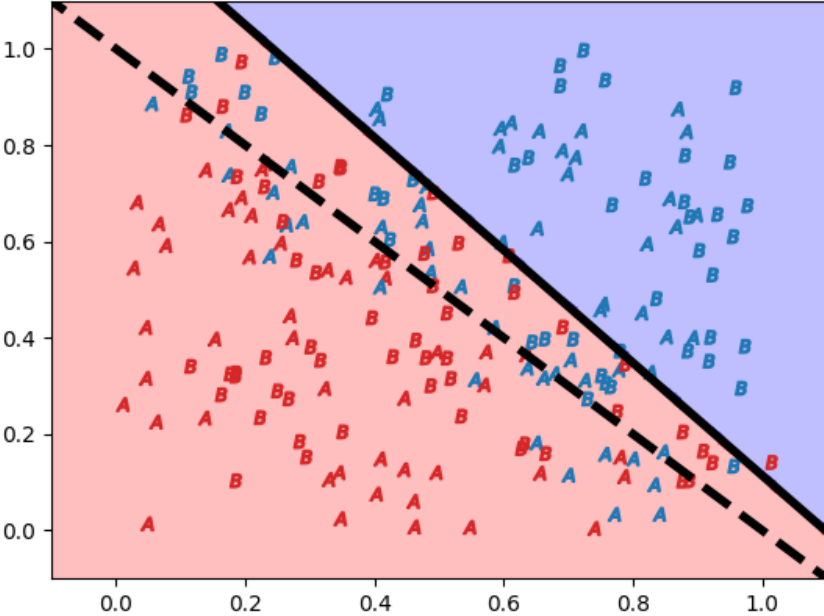}
        \caption{Robust Classifier (r=0.18)}
        \label{fig:num-syn-comp-Robust}
        \begin{tabular}{|c|c|c|}
            \hline
            \textbf{Diff:} & $Y=0, |S1-S0|$ & $Y=1, |S1-S0|$ \\
            \hline
            \hline
            Ind. & 0.025 & 0.025 \\
            \hline
            Sep. & 0.019 & 0.127 \\
            \hline
            Suff. & 0.142 & 0.038 \\
            \hline
        \end{tabular} \\
        \smallskip
        \textbf{Training Accuracy:} 73.5\% \\
        \textbf{Test Accuracy:} 73.0\%
    \end{subfigure}
    \caption{Comparative Analysis of Non-Robust and Robust Classifiers}
    \label{fig:num-syn-comp}
\end{figure}
We compare nonrobust training on this synthetic dataset with robust training (using the TRS method) and random perturbation for 11 different perturbation radii, with the radius increasing in 0.01 increments from 0.1 up to 0.2. These experiments were run with a total of 10 training epochs and a learning rate of 0.01 in the outer optimization problem. Four plots of fairness metric differences versus radius are shown in \cref{tab:num-syn-results}, as well as plots of the training and testing accuracy versus radius. 
For many radii in the lower end of the plotted range, many of the fairness differences are worse in the training data with robust training than with nonrobust. However, some fairness improvement can be seen with robust training. 
\par In all cases, at least two of the three fairness metrics show a downward trend for robust training. This suggests that while robust training may worsen fairness for a very small radius, fairness improvement is possible at more appropriate radii. At a radius of 0.18, all of the robust fairness differences are better than the corresponding nonrobust ones in the training data, although it does lead to a decrease in test accuracy from around $78\%$ to $73\%$ (\cref{fig:num-syn-comp}). 
The nonrobust classifier is visualized in \cref{fig:num-syn-comp-Nonrobust} versus the robust classifier in \cref{fig:num-syn-comp-Robust}. In \cref{fig:num-syn-comp-Robust}, robust optimization is improving fairness by raising the bar, giving a positive classification to only the most qualified individuals. It eliminates nearly all of the false positive $B$s that exist with the original dashed classifier and greatly increases the quantity of blue $B$s in the red region, equalizing false negative rates. Increasing the radius even further to 0.2 with 20 epochs of training, the robust classifier eventually classifies nothing positively. Our robust classifier is not helping the disadvantaged $A$s in the process of improving fairness – it is only hurting the advantaged $B$s. While this does not provide an indication of how robust training would work on every dataset, it does illustrate that robust optimization may improve fairness in an unexpected or unintended way.
\begin{figure}[t]
    \centering
     \begin{subfigure}[b]{.49\textwidth}
         \centering
         \includegraphics[width=\linewidth]{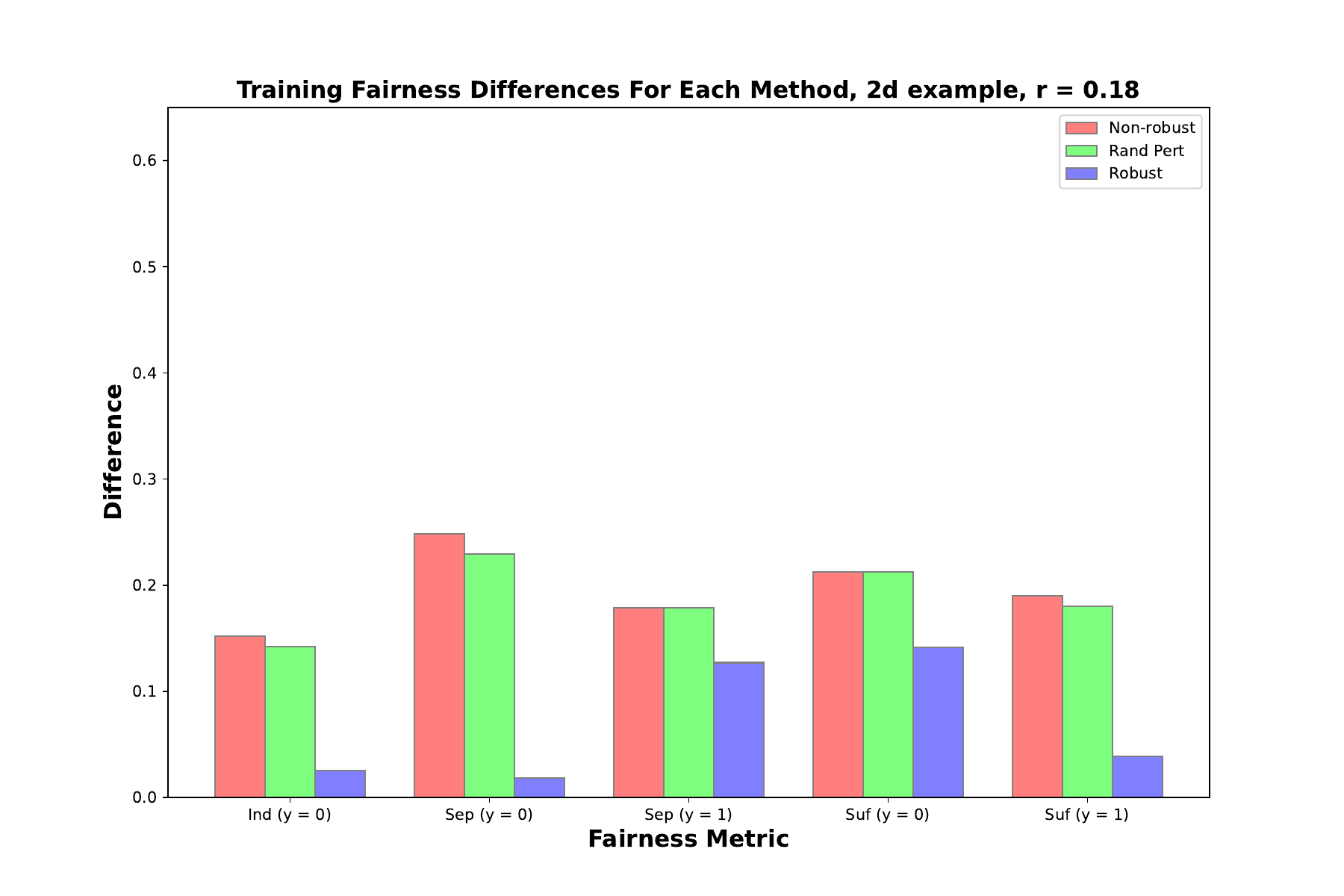}
         \caption{Training fairness differences}
         \label{fig:num-syn-bar-training}
     \end{subfigure}
     \hfill
     \begin{subfigure}[b]{0.49\textwidth}
         \centering
        \includegraphics[width=\linewidth]{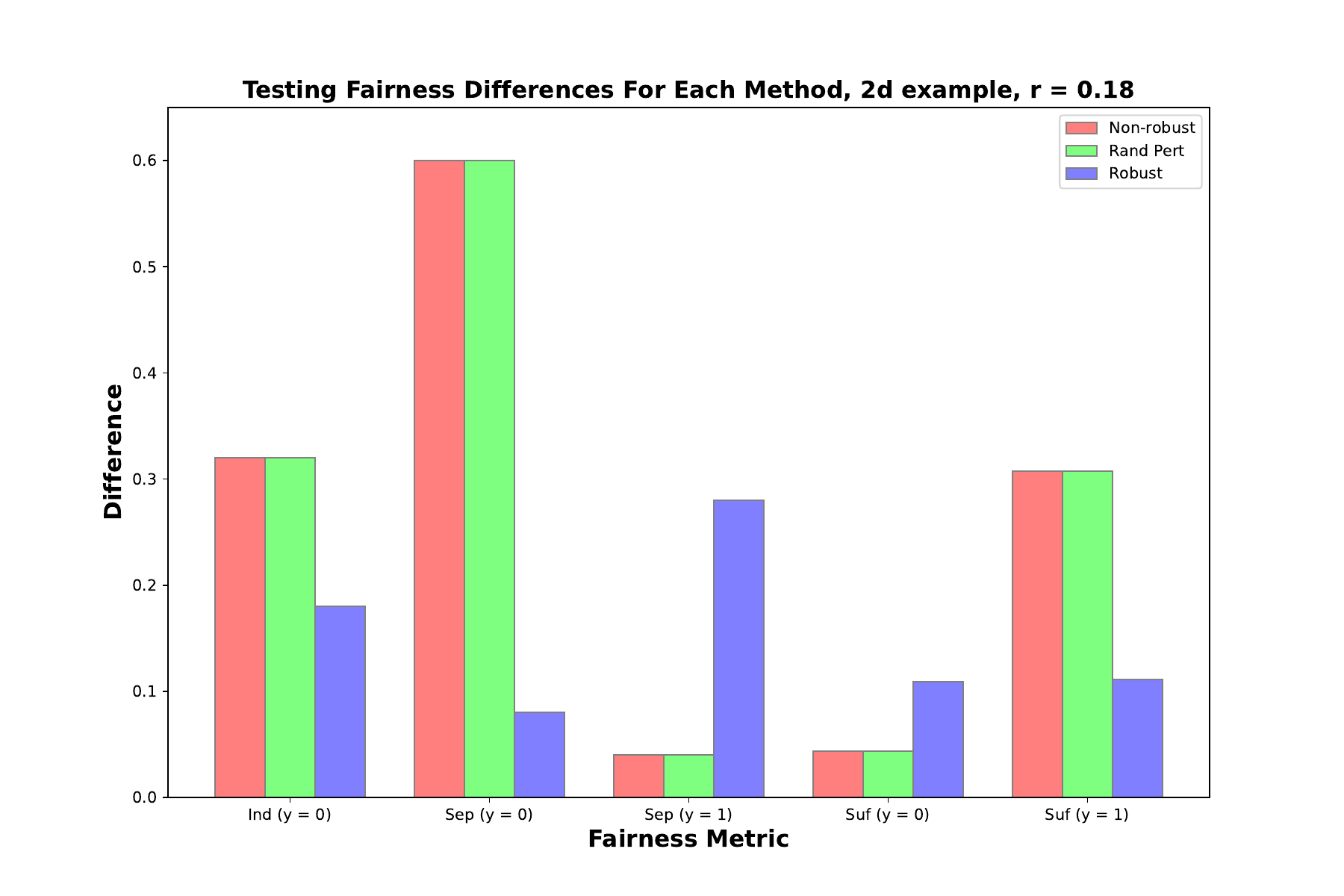}\\
         \caption{Test fairness differences}
         \label{fig:num-syn-bar-test}
     \end{subfigure}
     \caption{Fairness differences (r=0.18). Left bar is nonrobust, middle is random, right is robust.}
     \label{fig:num-syn-bar}
\end{figure} 
\par There is an advantage to solving the inner optimization problem well instead of just using random perturbations.  In \cref{tab:num-syn-results}, the fairness differences for random perturbations are either the same or stay close to the nonrobust differences. This stands in contrast to robust training, where fairness differences are initially high and then get significantly lower, surpassing nonrobust differences. \Cref{fig:num-syn-bar} also exhibits this advantage.

    \subsection{Adult Dataset}
    \label{sec:adult}
    \begin{figure}[t]
    \centering
    \includegraphics[width=0.4\linewidth]{Legend_for_graphs.pdf}
    
    \begin{subfigure}[t]{0.24\textwidth}
        \centering
        \includegraphics[width=\linewidth]{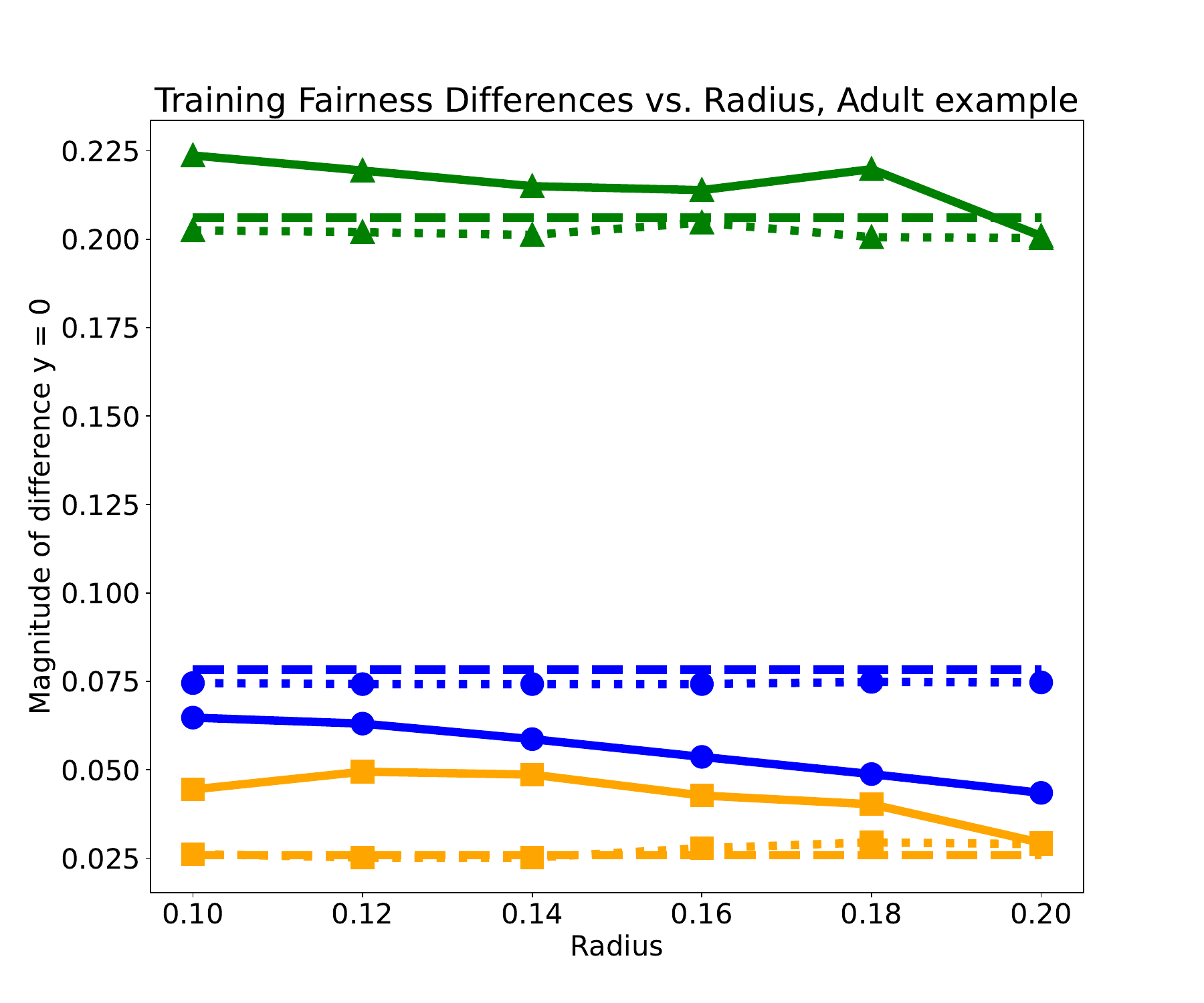}
        \caption{}
        \label{fig:num-AdulttrainingY0}
    \end{subfigure}
    \hfill 
    \begin{subfigure}[t]{0.24\textwidth} 
        \centering
        \includegraphics[width=\linewidth]{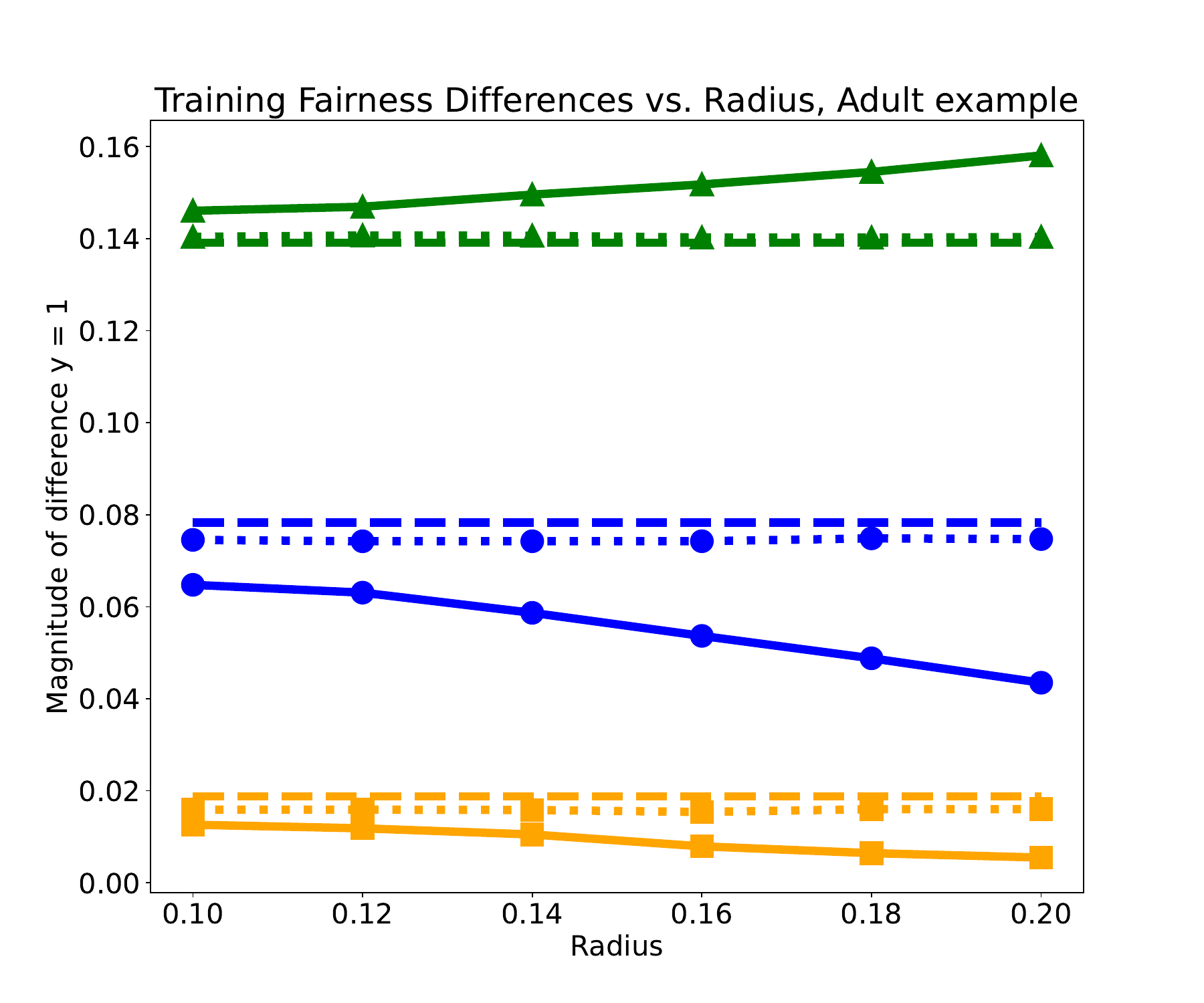}
        \caption{}
        \label{fig:num-AdulttrainingY1}
    \end{subfigure}
    \hfill 
    \begin{subfigure}[t]{0.24\textwidth} 
        \centering
        \includegraphics[width=\linewidth]{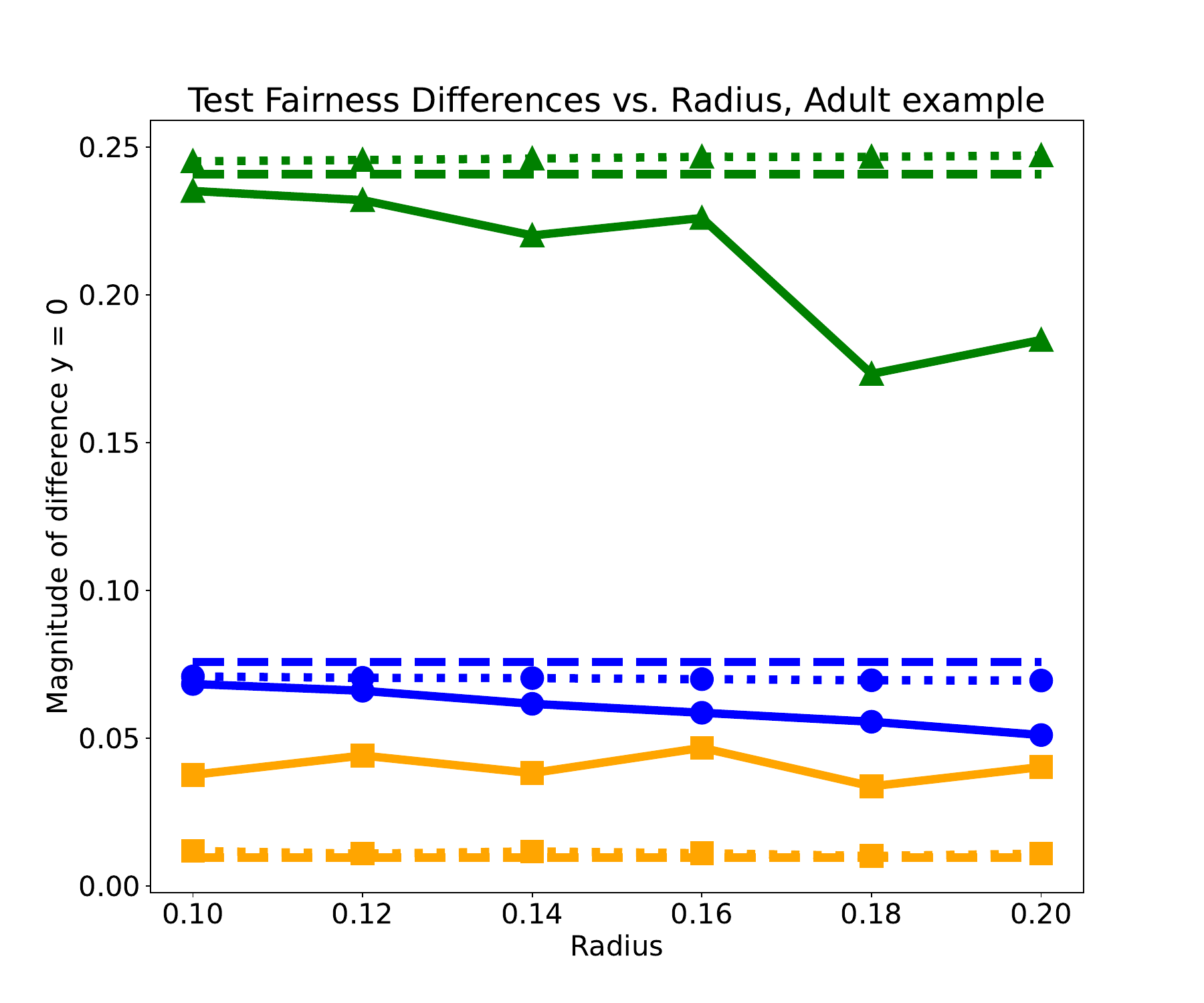}
        \caption{}
        \label{fig:num-AdulttestY0}
    \end{subfigure}
    \hfill 
    \begin{subfigure}[t]{0.24\textwidth} 
        \centering
        \includegraphics[width=\linewidth]{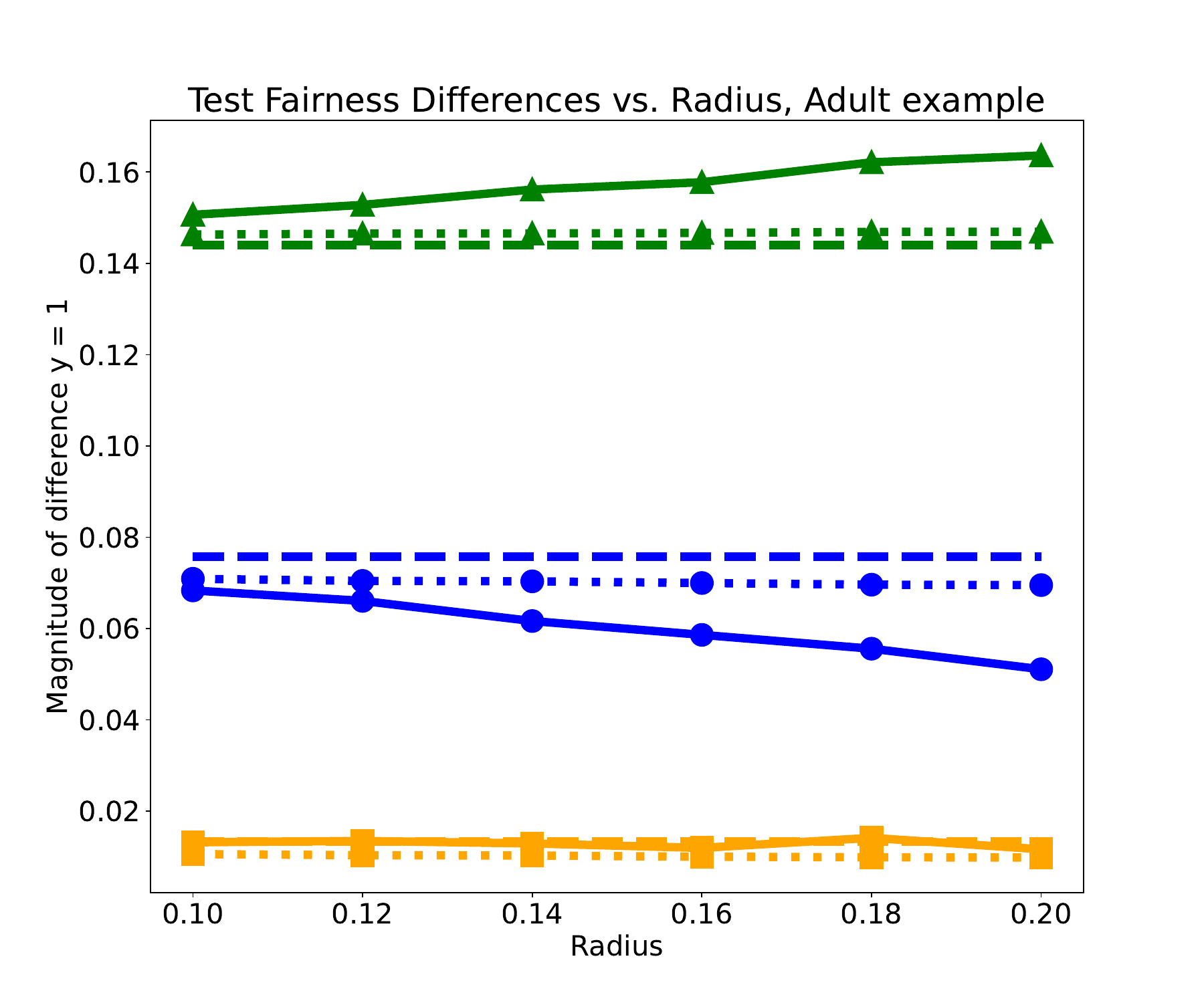}
        \caption{}
        \label{fig:num-AdulttestY1}
    \end{subfigure}
    \begin{subfigure}[b]{\textwidth}
        \centering
        \includegraphics[width=0.3\linewidth]{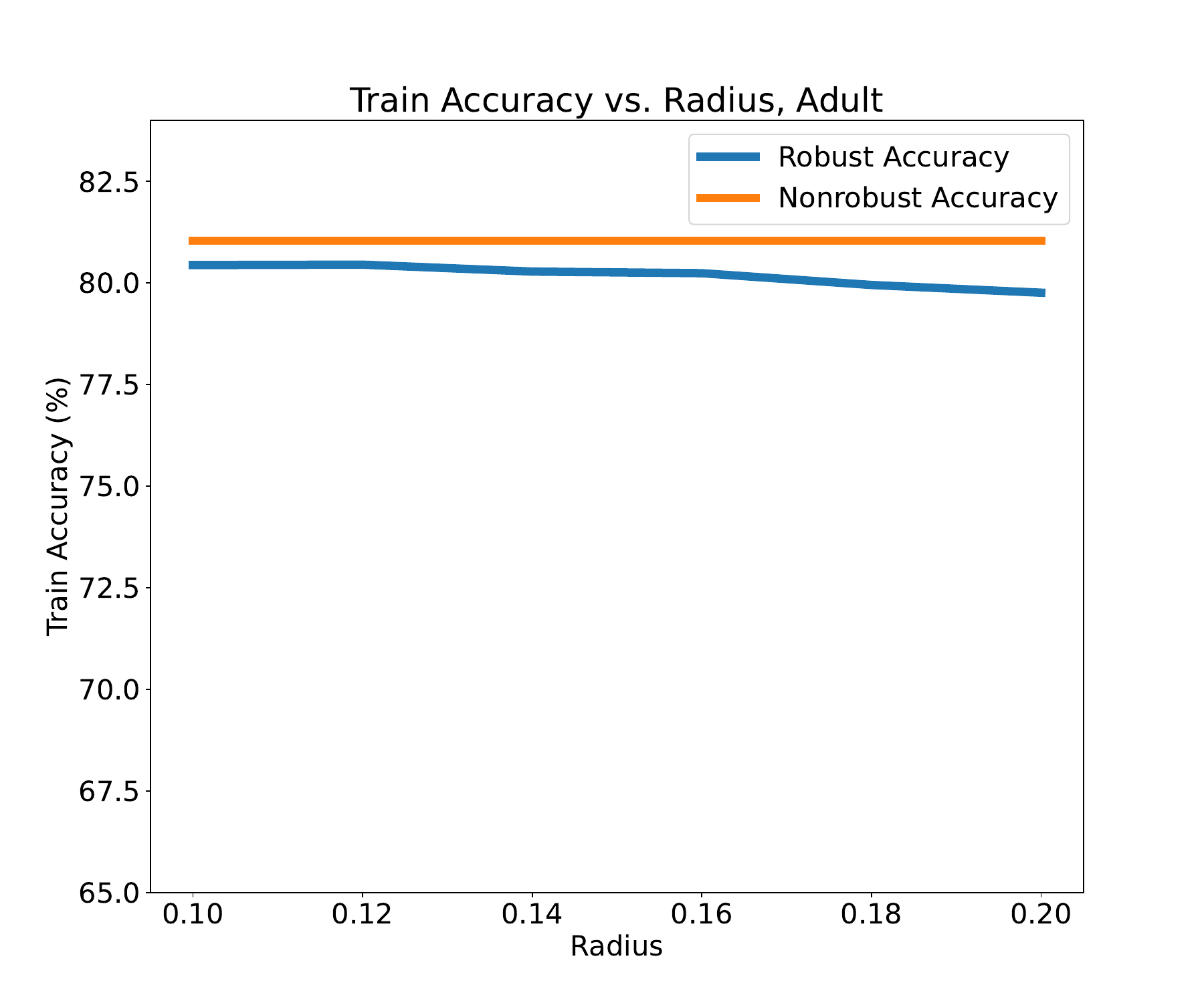}
        \includegraphics[width=0.3\linewidth]{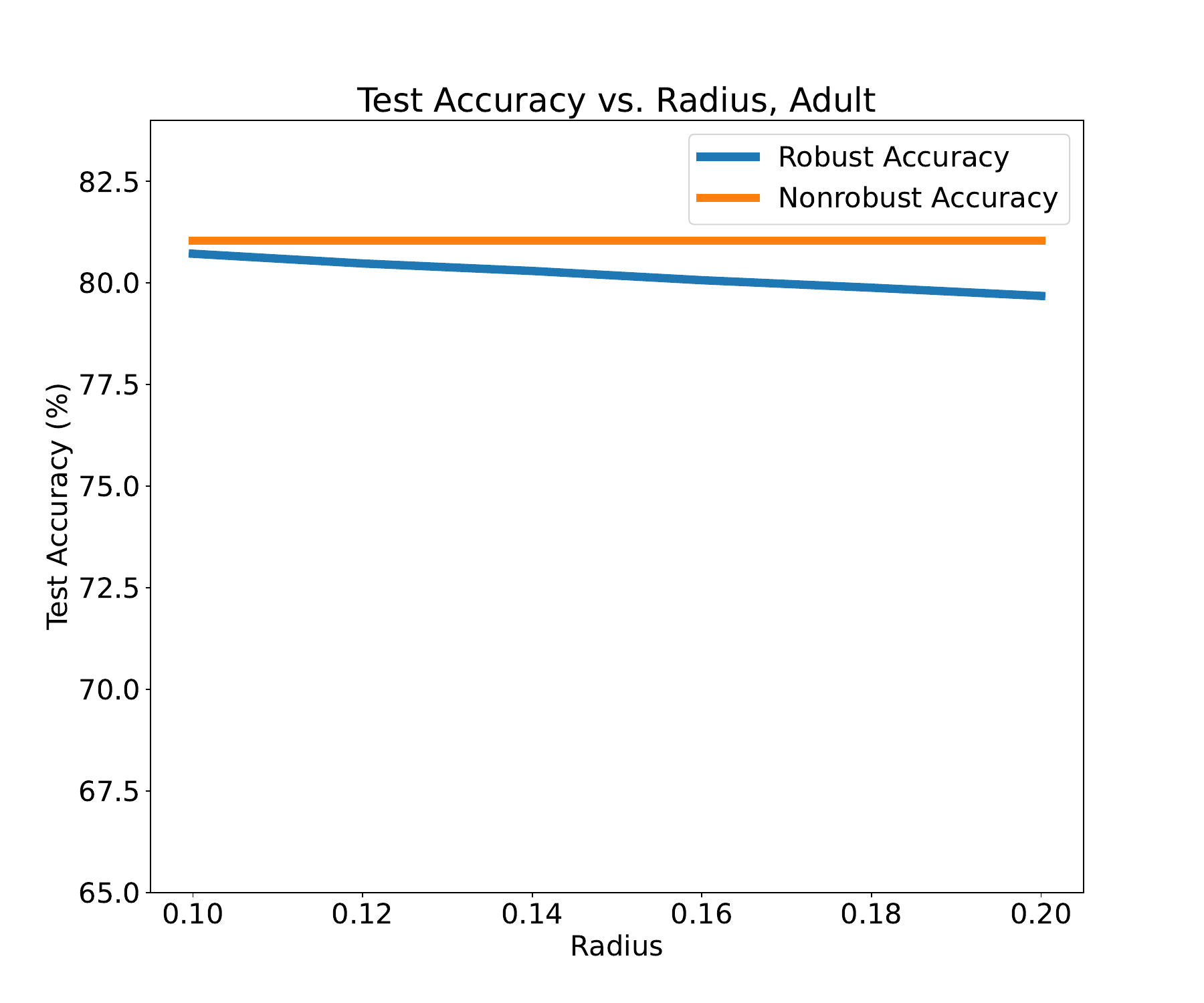}
        \caption{Adult robust training accuracy.}
        \label{fig:num-adult-accuracy}
    \end{subfigure}

    \caption{Fairness ((a)-(d)) and accuracy (e) trends in the adult dataset for nonrobust and robust training.}
    \label{fig:num-adult-results} 
\end{figure}

We also extended our numerical experiments to real-world datasets. The Adult dataset \cite{Le_Quy_2022} consists of demographic data about individuals that are used to classify whether their annual income is more than $\$50,000$. Note that the dataset predominantly consists of white males in the age range of 25-60. It contains 48,842 instances and each instance is described using 15 attributes. We want to look at the 5 continuous numerical attributes (age, education-num, capital-gain, capital-loss, hours-per-week) for analysis. The income (salary) data is converted into binary form (1 for $ \le50k$, 0 for $>50K$), and the protected attribute can be sex or race.

Unlike our synthetic data, the adult example yielded mixed results in terms of fairness improvement. For the training data, \cref{fig:num-AdulttrainingY0} and \cref{fig:num-AdulttrainingY1} show that only three out of the six differences were measured to be better with robust training. There was a similar result for the test data as seen in \cref{fig:num-AdulttestY0} and \cref{fig:num-AdulttestY1}. Despite only having a 50\% improvement rate, the majority of the fairness metrics exhibit a downward trend, and when there is an improvement robust optimization outperforms random perturbation significantly. The expected accuracy-robustness trade-off is present (\cref{fig:num-adult-accuracy}), with both the training and test robust accuracy decreasing with increasing radii. However, unlike in the synthetic dataset, the decay appears to be linear and does not spike at certain radii, and does not yield better accuracy for smaller radii. Overall, there is still reasonable case for improving fairness metrics at the expense of classifier accuracy. 

    \subsection{LSAT Data}
    \label{sec:lsat}
    Another extension to a real-world dataset comes from the Law School Admissions Council (LSAC) \cite{Wightman1998LSACNL}. This dataset was collected to explore the reasons behind low bar passage rates among racial and ethnic minorities. We train our classifier to predict whether or not a student will pass the bar, based on their Law School Admission Test (LSAT) score and undergraduate GPA. We are using GPA and LSAT scores because they are the strongest predictors for passing the bar examination. Our primary interest lies in examining five key features of the dataset: the bar exam pass/fail prediction made by a DNN, the gender of the student, their LSAT score, the true bar exam pass/fail value for the student, and their race. For the purpose of our experiment, the race feature is made binary to indicate a student as either white or non-white, which is used as the sensitive attribute.
\begin{figure}[t]
    \centering
    \includegraphics[width=0.4\linewidth]{Legend_for_graphs.pdf}
    
    \begin{subfigure}[t]{0.24\textwidth} 
        \centering
        \includegraphics[width=\linewidth]{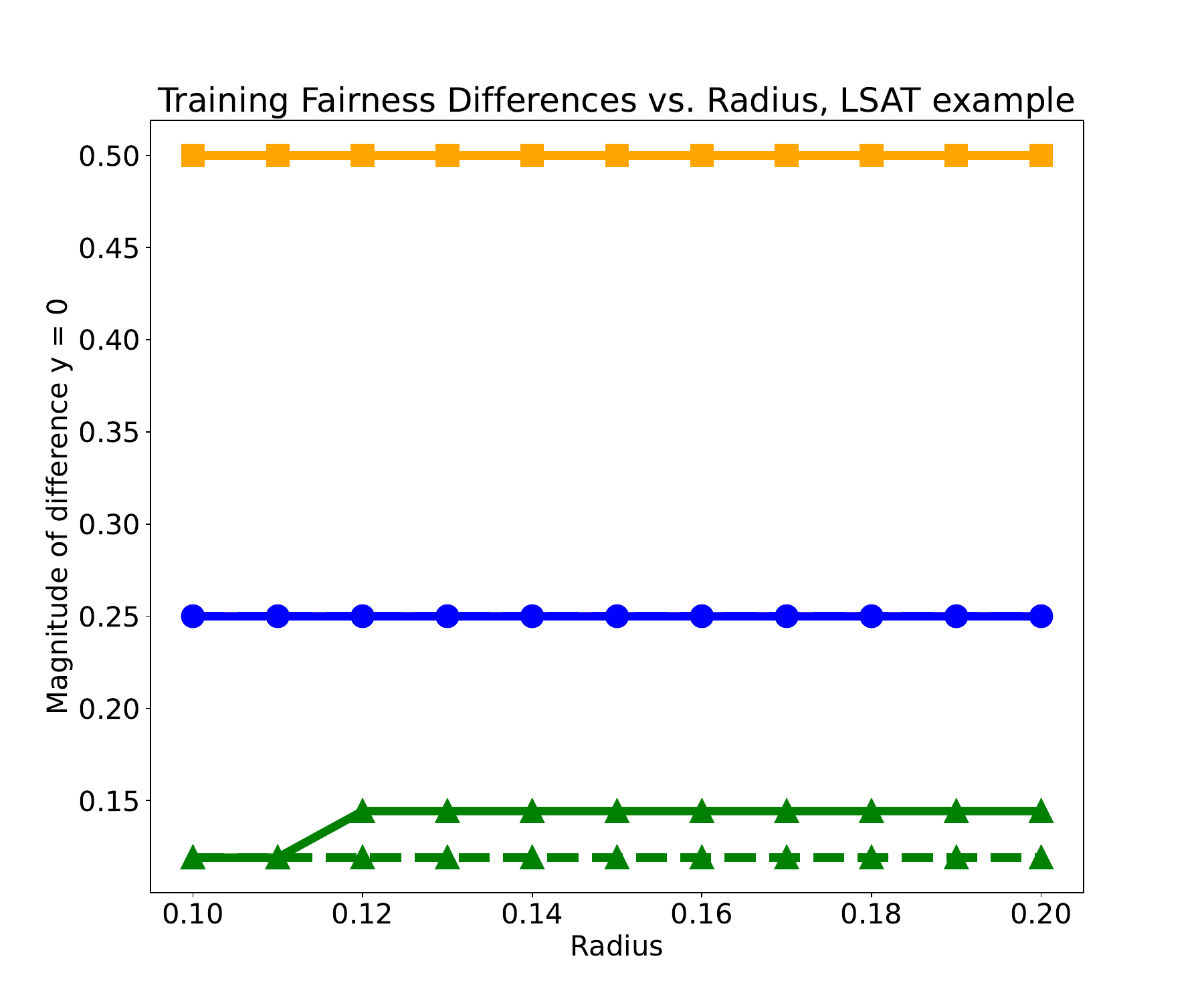}
        \caption{}
        \label{fig:num-lsat-LSATtrainingY0}
    \end{subfigure}
    \hfill 
    \begin{subfigure}[t]{0.24\textwidth} 
        \centering
        \includegraphics[width=\linewidth]{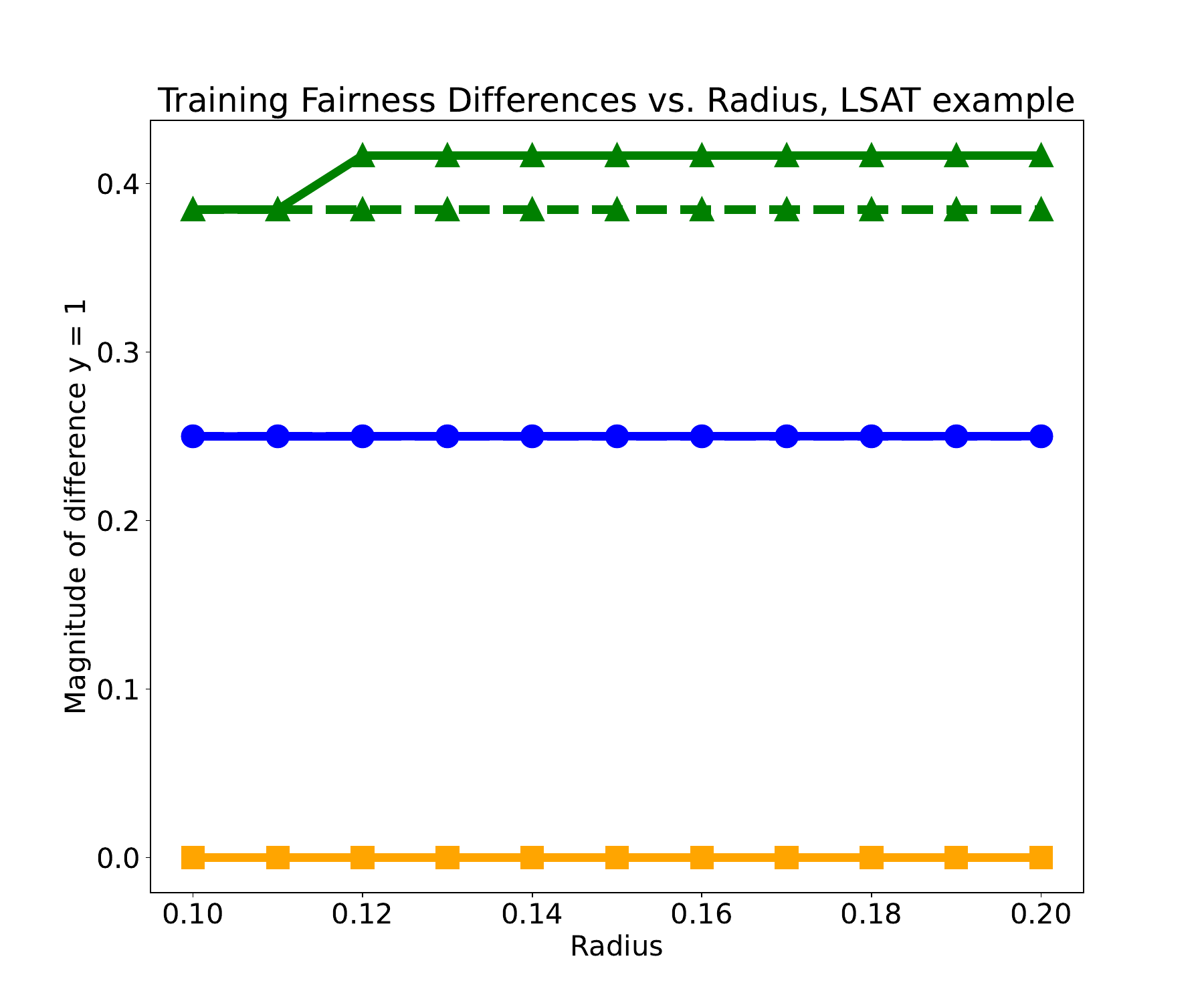}\\
        \caption{}
        \label{fig:num-lsat-LSATtrainingY1}
    \end{subfigure}
    \hfill 
    \begin{subfigure}[t]{0.24\textwidth} 
        \centering
        \includegraphics[width=\linewidth]{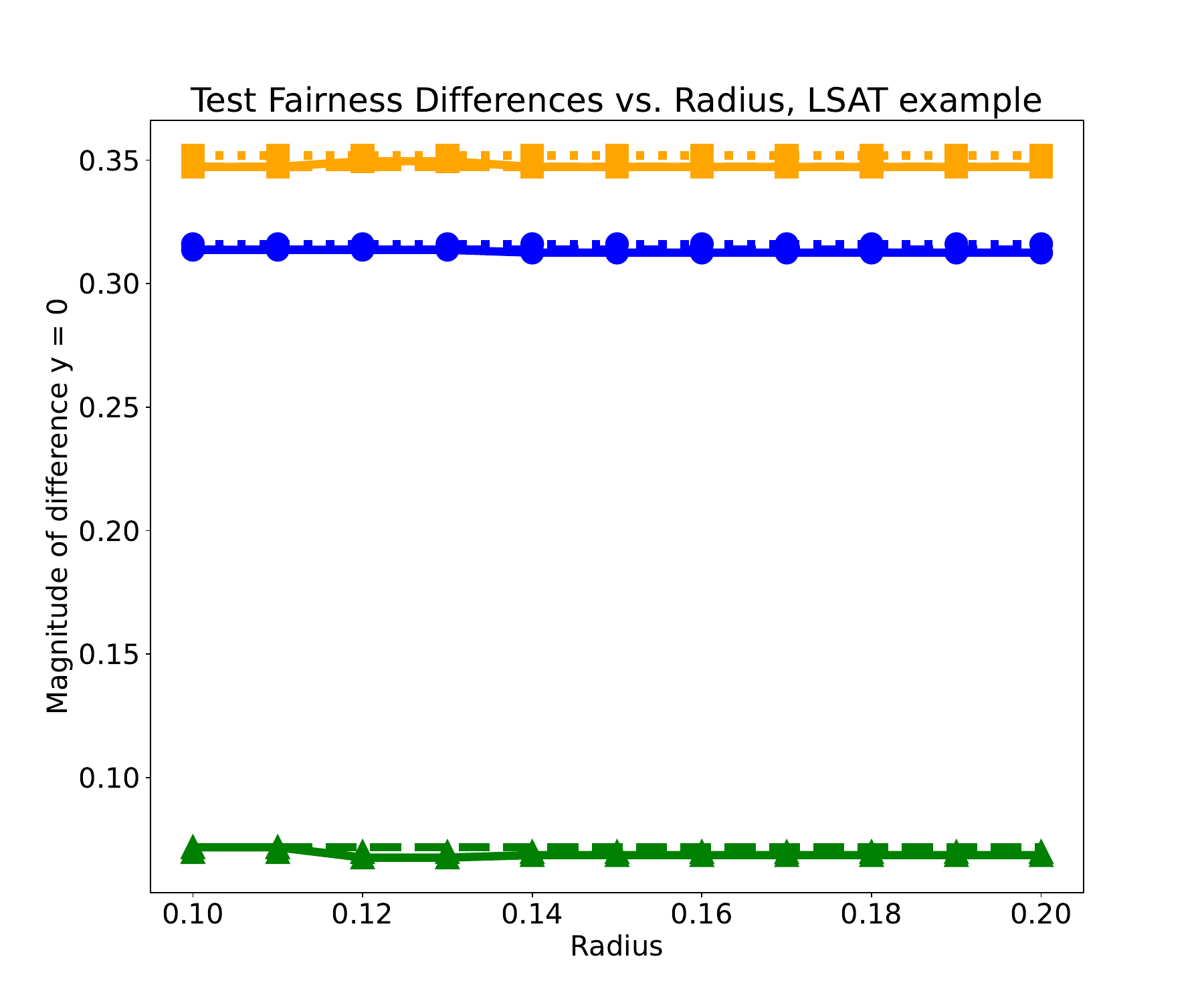}
        \caption{}
        \label{fig:num-lsat-LSATtestY0}
    \end{subfigure}
    \hfill
    \begin{subfigure}[t]{0.24\textwidth} 
        \centering
        \includegraphics[width=\linewidth]{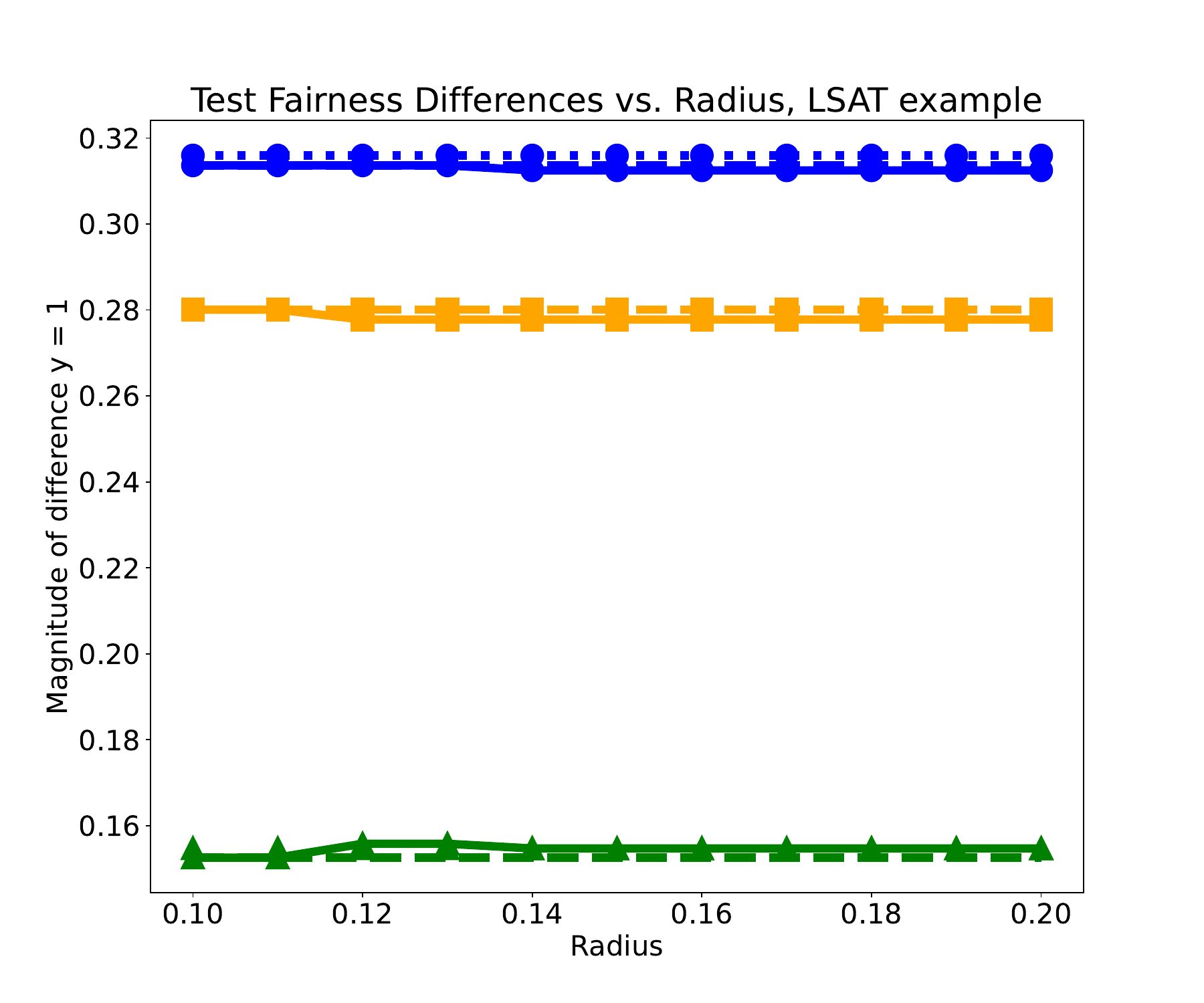}
        \caption{}
        \label{fig:num-lsat-LSATtestY1}
    \end{subfigure}
    \begin{subfigure}[b]{\textwidth}
        \centering
        \includegraphics[width=0.3\linewidth]{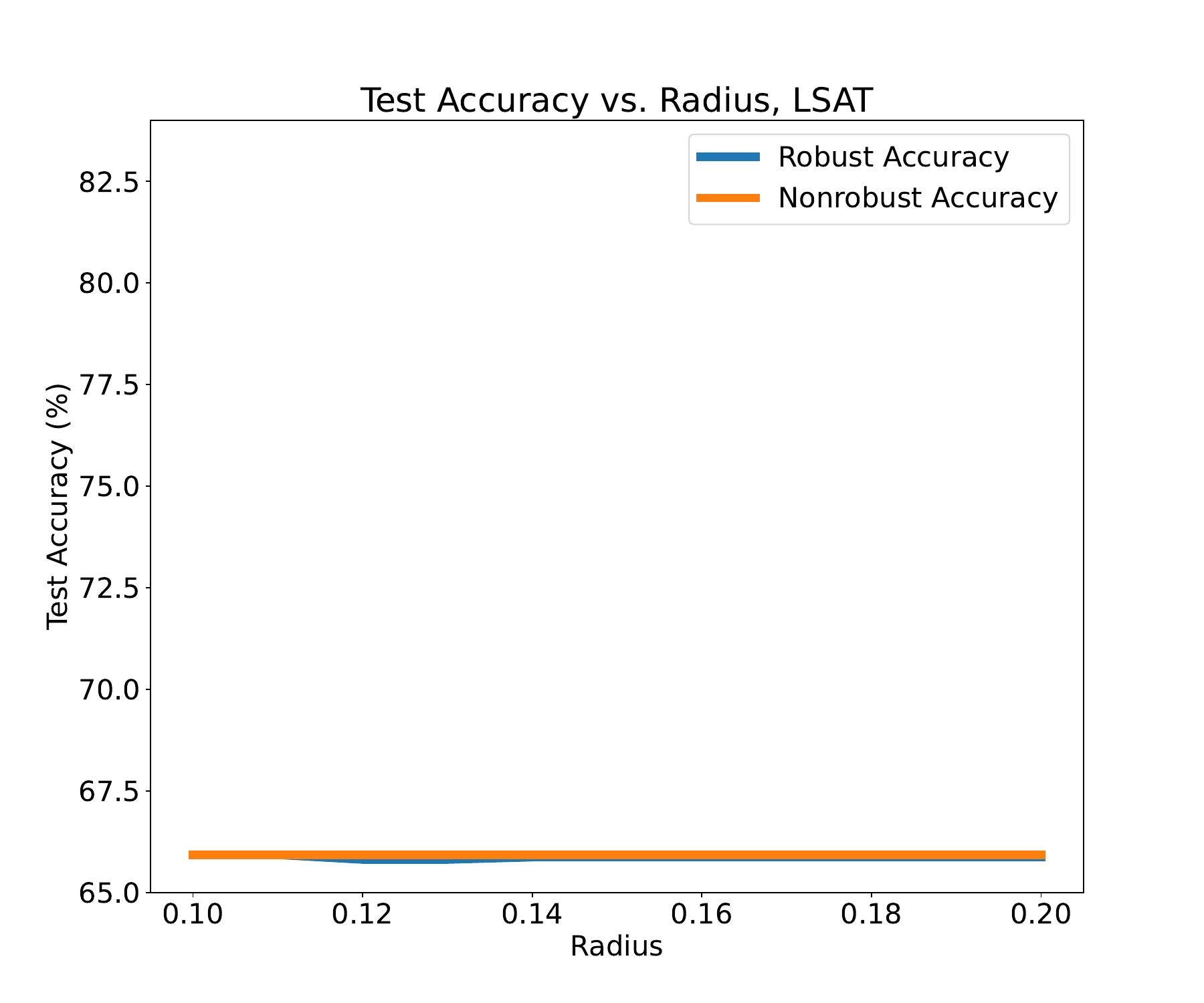}
        \includegraphics[width=0.3\linewidth]{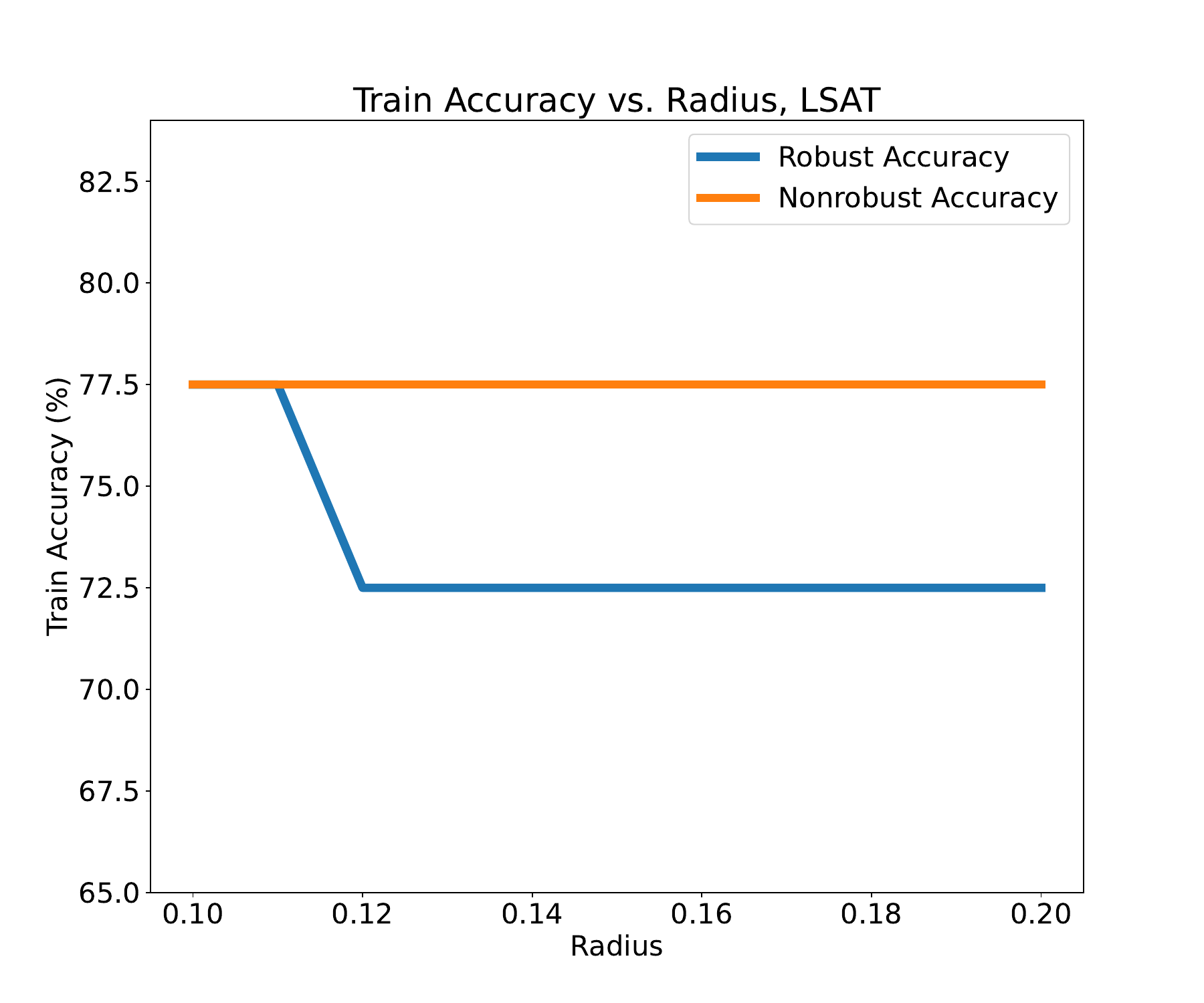}
        \caption{LSAT robust training accuracy.}
        \label{fig:num-lsat-accuracy}
    \end{subfigure}

    \caption{Fairness ((a)-(d)) and accuracy (e) trends in the LSAT dataset for nonrobust and robust training.}
    \label{fig:num-lsat-fairness} 
\end{figure}

Unlike the two previous examples, there is not a lot of fluctuation with LSAT robust results (\cref{fig:num-lsat-fairness}). Surprisingly, robust optimization seems not to deviate from nonrobust training. As we have seen, using a random perturbation yields fairness results that are very similar to the results of non-robust training. For this dataset, robust optimization also performs very similarly to just picking a random perturbation. This is especially the case with the independence and separation metrics. The only two notable deviations come from the sufficiency metric in \cref{fig:num-lsat-LSATtrainingY0} and \cref{fig:num-lsat-LSATtrainingY1} of the training dataset. This might be attributable to our definion of the sensitive attribute as white vs nonwhite, which creates a very small dataset due to the dominance of white individuals in the original dataset. For this specific example, robust optimization does not improve fairness, and in fact performs worse since it loses accuracy when yielding the same fairness results.

    \subsection{Efficiency Comparison}
    \label{sec:efficiency}
    \begin{table}[t]
  \centering
  \caption{Average Epoch Times. For each dataset, the first three rows show the average epoch times for each of the three robust optimization approaches, where  a lower value indicates faster computational performance. The fourth row shows the ratio of PGD to TRS time, where a ratio greater than 1 indicates that the TRS approach was faster than the PGD approach for the given radius. The gray values highlight the minimum ratio of PGD to TRS time over all radii, while the yellow values highlight the maximum ratio.}
  \begin{tabular}{|c|c||c|c|c|c|c|c|}
    \hline
    \textbf{} & \textbf{Radii} & .10 & .12 & .14 & .16 & .18 & .20 \\
    \hline
    \multirow{4}{*}{\rotatebox[origin=c]{90}{\textbf{Synthetic}}} & \textbf{PGD} & 2.680 & 3.597 & 4.228 & 4.486 & 5.061 & 5.080  \\
    \cline{2-8}
    & \textbf{TRS} & 1.945 & 1.875 & 1.806 & 1.891 & 1.742 & 1.752 \\
    \cline{2-8}
    & \textbf{RND} & 0.0387 & 0.0366 & 0.0387 & 0.0389 & 0.0387 & 0.0375 \\
    \cline{2-8}
    & \textbf{PGD/TRS} & \sethlcolor{lightgray}\hl{1.377} & 1.919 & 2.340 & 2.373 & \hl{2.904} & 2.900 \\
    \hline
    \hline
    \multirow{4}{*}{\rotatebox[origin=c]{90}{\textbf{Adult}}} & \textbf{PGD} &512.970 &  593.173 & 697.399 & 1146.856 & 1689.761 & 1854.932 \\
    \cline{2-8}
    & \textbf{TRS} & 60.372 & 61.557 & 57.723 & 58.707 & 57.492 & 59.061 \\
    \cline{2-8}
    & \textbf{RND} & 0.0947 & 0.101 & 0.0972 & 0.0919 & 0.0974 & 0.0939 \\
    \cline{2-8}
    & \textbf{PGD/TRS} & \sethlcolor{lightgray}\hl{8.497} & 9.636 & 12.082 & 19.535 & 29.391 & \hl{31.407} \\
    \hline
    \hline
    \multirow{4}{*}{\rotatebox[origin=c]{90}{\textbf{LSAT}}} & \textbf{PGD} & 1.129 & 1.559 & 2.844 & 3.575 & 3.347 &  2.752\\
    \cline{2-8}
    & \textbf{TRS} & 0.396 & 0.396 &  0.421 & 0.371 &  0.414 & 0.385\\
    \cline{2-8}
    & \textbf{RND} & 0.0107 & 0.0123 & 0.0154 & 0.0122 & 0.0162 & 0.0104 \\
    \cline{2-8}
    & \textbf{PGD/TRS} & \sethlcolor{lightgray} \hl{2.852} & 3.936 & 6.762 & \hl{9.639} & 8.094 & 7.150 \\
    \hline
  \end{tabular}
  \label{tab:num-eff-epoch times}
\end{table}

We gathered time data to see if the TRS method converged faster than PGD on our datasets. On each of the three datasets above - synthetic, Adult, and LSAT - we computed for each radius the average epoch time elapsed for TRS, PGD, and random perturbation. To compare the speed of the TRS method and PGD, we examine the ratio of the average PGD epoch time to the average TRS epoch time, looking at the extreme values of this ratio to get a range of how much faster the trust region subproblem was than PGD across all radii. The results are shown in \cref{tab:num-eff-epoch times}.

Random perturbation is the fastest adversarial training method in all three datasets. This is expected, as it does not actually solve the optimization problem; its only computation task is generating a random vector and rescaling it. It is also noteworthy that using the TRS method consistently is computationally faster than using PGD. Over all radii shown, training with TRS is between 1.4 and 2.9 times faster than PGD in the synthetic dataset, between 8.5 and 31.4 times faster in Adult, and between 2.9 and 9.6 times faster in LSAT. The very short average epoch times for the LSAT dataset are due to the significantly smaller scale of the input data. All of the smallest factors of time improvement of TRS relative to PGD (highlighted in gray) are greater than 1 suggesting that the trust region subproblem has a consistent advantage over PGD in computational speed. 

Looking at the PGD/TRS ratios, the factor of improvement that TRS has in computational time over PGD appears to be higher in the real-world datasets than in the synthetic dataset. The real-world datasets, and especially Adult, are trained on larger amounts of data, so the advantage of TRS over PGD seems to scale with the size of the dataset. This advantage of the trust region subproblem also improves with larger perturbation radii. In particular, the minimum factor of improvement (gray) always occurs with the smallest radius, and the maximum factor of improvement (yellow) always occurs with one of the three largest radii. 
    
\section{Conclusion}
\label{sec:conclusion}
In our affine linear model setup, we were able to see improvement in fairness by using robust optimization. In the synthetic dataset, whenever there was an improvement, the gain was a significant reduction in fairness difference magnitudes (which are ideally zero). In our numerical experiments extending to real-world datasets, we have shown that robust training performs similarly to non-robust training even in the worst-case scenario (LSAT dataset). Across all three datasets, the accuracy of robust optimization decreased as the radius increased, the majority of the fairness metrics displayed a downward trend as the perturbation radius increased, and when fairness improved with robust training, precise solutions to the inner optimization problem outperformed randomly selected solutions. Furthermore, we were able to quantify the fact that, with the help of \texttt{hessQuik}, using second-order information is much faster for solving our class of optimization problem. 

We acknowledge that while we were able to achieve positive results with our experiments in both synthetic and real-world datasets, there are a few mathematical limitations to our results that prevent generalization to higher-dimensional applications. We used a neural network in our training with only one hidden layer, our experiments were conducted using a linear and binary classifier, and our sensitive attribute was binary. This motivates future exploration of extending our approach to deeper neural networks, multinomial classification, and other fairness metrics relevant to those cases. It may help to improve fairness even further to introduce a regularization term to our approach to penalize violations of our fairness metrics, which is another avenue for further work. There limitations of our implementation. For PGD, we used an arbitrary step size instead of varying the step size as training proceeds. Additionally, we did not solve our inner optimization problems in parallel. Parallelizing the computations for our inner optimization problem could provide a significant reduction in overall computation time. 

Despite these limitations, this work demonstrates initial promise for the ability of robust training to bring about fairness improvement in machine learning models, and motivates further research on similar methodologies.

\section*{Acknowledgments}
This work was supported by NSF award DMS-2051019 and was completed during the ``Computational Mathematics for Data Science'' REU/RET program in the summer of 2023. We would like to thank Dr. Elizabeth Newman, our mentor, and the rest of the faculty who participated in the program for their feedback and support.

\bibliographystyle{siamplain}
\bibliography{main}

\end{document}


\maketitle

\section{A detailed example}

Here we include some equations and theorem-like environments to show
how these are labeled in a supplement and can be referenced from the
main text.
Consider the following equation:
\begin{equation}
  \label{eq:suppa}
  a^2 + b^2 = c^2.
\end{equation}
You can also reference equations such as \cref{eq:matrices,eq:bb} 
from the main article in this supplement.

\lipsum[100-101]

\begin{theorem}
An example theorem.
\end{theorem}

\lipsum[102]
 
\begin{lemma}
An example lemma.
\end{lemma}

\lipsum[103-105]

Here is an example citation: \cite{KoMa14}.

\section[Proof of Thm]{Proof of \cref{thm:bigthm}}
\label{sec:proof}

\lipsum[106-112]

\section{Additional experimental results}
\Cref{tab:foo} shows additional
supporting evidence. 

\begin{table}[htbp]
\footnotesize
  \caption{Example table.}  \label{tab:smfoo}
\begin{center}
  \begin{tabular}{|c|c|c|} \hline
   Species & \bf Mean & \bf Std.~Dev. \\ \hline
    1 & 3.4 & 1.2 \\
    2 & 5.4 & 0.6 \\ \hline
  \end{tabular}
\end{center}
\end{table}

\bibliographystyle{siamplain}
\bibliography{references}